\documentclass[10pt,twocolumn,letterpaper]{article}
\usepackage{cvpr}      % To produce the REVIEW version
\usepackage{graphicx}
\usepackage{amsmath}
\usepackage{amssymb}
\usepackage{lipsum}
\usepackage{booktabs}
\usepackage[symbol]{footmisc}
\usepackage[accsupp]{axessibility}
\usepackage[pagebackref,breaklinks,colorlinks]{hyperref}
\usepackage[capitalize]{cleveref}
\crefname{section}{Sec.}{Secs.}
\Crefname{section}{Section}{Sections}
\Crefname{table}{Table}{Tables}
\crefname{table}{Tab.}{Tabs.}
\def\cvprPaperID{5037} % *** Enter the CVPR Paper ID here
\def\confName{CVPR}
\def\confYear{2023}
\usepackage{cite}
\usepackage[utf8]{inputenc} % allow utf-8 input
\usepackage[T1]{fontenc}    % use 8-bit T1 fonts
\usepackage{hyperref}       % hyperlinks
\usepackage{url}            % simple URL typesetting
\usepackage{booktabs}       % professional-quality tables
\usepackage{float}
\usepackage{amsfonts}       % blackboard math symbols
\usepackage{nicefrac}       % compact symbols for 1/2, etc.
\usepackage{microtype}      % microtypography
\usepackage{xcolor}         % colors
\usepackage{hyperref}       % hyperlinks
\usepackage{url}            % simple URL typesetting
\usepackage{booktabs}       % professional-quality tables
\usepackage{amsfonts}       % blackboard math symbols
\usepackage{nicefrac}       % compact symbols for 1/2, etc.
\usepackage{microtype}      % microtypography
\usepackage{enumerate}
\usepackage{cuted}
\usepackage{subcaption}
\usepackage{booktabs}
\usepackage{multirow}
\usepackage{amsmath}
\usepackage{amssymb}
\usepackage{wrapfig}
\usepackage{hyperref}
\usepackage{pifont}
\usepackage{color}

\usepackage[ruled]{algorithm2e}
\usepackage{algorithmic}

\newtheorem{definition}{Definition}[section]

\usepackage{array}
\newcolumntype{L}[1]{>{\raggedright\let\newline\\\arraybackslash\hspace{0pt}}m{#1}}
\usepackage{paralist}

\usepackage{amsmath,amsfonts,bm}

\def\eqref#1{equation~\ref{#1}}
\def\1{\bm{1}}

\DeclareMathAlphabet{\mathsfit}{\encodingdefault}{\sfdefault}{m}{sl}
\SetMathAlphabet{\mathsfit}{bold}{\encodingdefault}{\sfdefault}{bx}{n}

\def\sX{{\mathbb{X}}}

\usepackage{hyperref}
\usepackage{url}

\title{Exploring and Exploiting Uncertainty for Incomplete Multi-View Classification}
\author{Mengyao Xie\thanks{Equal contributions.}, Zongbo Han\footnotemark[1], Changqing Zhang\thanks{Corresponding author.}, Yichen Bai, Qinghua Hu\\
College of Intelligence and Computing, Tianjin University\\
\texttt{\small zhangchangqing@tju.edu.cn}}

\begin{document}

\maketitle

%%%%%%%%% ABSTRACT
\begin{abstract}
Classifying incomplete multi-view data is inevitable since arbitrary view missing widely exists in real-world applications. Although great progress has been achieved, existing incomplete multi-view methods are still difficult to obtain a trustworthy prediction due to the relatively high uncertainty nature of missing views. First, the missing view is of high uncertainty, and thus it is not reasonable to provide a single deterministic imputation. Second, the quality of the imputed data itself is of high uncertainty. To explore and exploit the uncertainty, we propose an Uncertainty-induced Incomplete Multi-View Data Classification (UIMC) model to classify the incomplete multi-view data under a stable and reliable framework. We construct a distribution and sample multiple times to characterize the uncertainty of missing views, and adaptively utilize them according to the sampling quality. Accordingly, the proposed method realizes more perceivable imputation and controllable fusion. Specifically, we model each missing data with a distribution conditioning on the available views and thus introducing uncertainty. Then an evidence-based fusion strategy is employed to guarantee the trustworthy integration of the imputed views. Extensive experiments are conducted on multiple benchmark data sets and our method establishes a state-of-the-art performance in terms of both performance and trustworthiness. 
% Multiple modalities can yield more comprehensive information for classification task than single modality. Partial multimodal classification problem(PMCF) aims to obtain accuracy as high as possible utilizing incomplete multimodal data. Although significant progress has been achieved by existing PMCF methods, there still exist disadvantages i) neglecting missing modalities causes correlation among modalities can not be explored and poor performance under high missing rates and ii) some imputation strategies may introduce noise. In view of the shortcomings, we propose UIMC to impute missing data utilizing consistency among modalities and choose imputed data beneficial to classification through an uncertainty-guided weight strategy. Then Subjective logic and Dempster’s Combination Rule are introduced to classify imputed complete multimodal data credibly. We conduct extensive experiments to demonstrate that our UIMC can achieve optimal performance basically and is more robust to missing degree in comparison with state-of-the-art PMCF methods.
\end{abstract}
%%%%%%%%% BODY TEXT
\section{Introduction}
\label{sec:intro}
Learning from multiple complementary views has the potential to yield more generalizable models. Benefiting from the power of deep learning, multi-view learning has further exhibited remarkable benefits against the single-view paradigm in clustering~\cite{clu_ensemble, clu_matrix, clu_reliable}, classification~\cite{class_consistent, class_embedded} and representation learning~\cite{repre_survey, repre_flexible}. However, real-world data are usually incomplete. For instance, in the medical field, patients with same condition may choose different medical examinations producing incomplete/unaligned multi-view data; similarly, sensors in cars at times may be out of order and thus only part of information can be collected. Flexible and trustworthy utilization of incomplete multi-view data is still very challenging due to the high uncertainty of missing issues.

For the task of incomplete multi-view classification (IMVC), there have been plenty of studies which could be roughly categorized into two main lines. The methods~\cite{DeepIMV, CPM} only use available views without imputation to conduct classification. While the other line~\cite{MVAE, MIWAE, impute_ADMM, impute_adversarial} reconstructs missing data based on deep learning methods such as autoencoder~\cite{AE1, AE2} or generative adversarial network (GAN)~\cite{GAN}, and then utilizes the imputed complete data for classification. Although significant progress has been achieved by existing IMVC methods, there are still limitations:
(1) Methods that simply neglect missing views are usually ineffective especially under high missing rate due to the limitation in exploring the correlation among views; 
(2) Methods that impute missing data based on deep learning methods are short in interpretability and the deterministic imputation way fails to characterize the uncertainty of missing resulting in unstable classification;
(3) Few IMVC methods can handle multi-view data with complex missing patterns especially for the data with more than two views, which makes these methods inflexible.

In view of above limitations, we propose a simple yet effective, stable and flexible incomplete multi-view classification model. First, the proposed model characterizes the uncertainty of each missing view by imputing a distribution instead of a deterministic value. The necessity of characterizing the uncertainty for missing data on single view has been well recognized in recent works~\cite{supMIWAE, impute_uncertainty}. Second, we conduct sampling multiple times from the above distribution and each one is combined with the observed views to form multiple completed multi-view samples. The quality of the sampled data is of high uncertainty, and thus we adaptively integrate them according to their quality on the single view and multi-view fusion. For single view, the uncertainty of the low-quality sampled data tends to be large and should not affect the learning of other views. Therefore, we construct an evidence-based classifier for each view to obtain the opinion including subjective probabilities and uncertainty masses. While for multi-view fusion, the opinions from multiple views are integrated based on DS rule, which ensures the trustworthy utilization of arbitrary-quality views. 
%The quality of sampled data is of high uncertainty, and thus we adaptively utilize them according to the sampling quality in two stages. In the training stage, the losses of the low-quality sampled data tend to be large and should not affect the learning of other views. Therefore, we dynamically weight the loss for different sampled data. Second, in the fusion stage, the opinions from multiple modalities are integrated based on DS rule, where ensures the trustworthy utilization of arbitrary-quality views. 
The main contributions of this work are summarized as follows:
\begin{itemize}
    \item We propose an exploration-and-exploitation strategy for classifying incomplete multi-view data by characterizing the uncertainty of missing data, which promotes both effectiveness and trustworthiness in utilizing imputed data. To the best of our knowledge, the proposed UIMC is the first work asserting uncertainty in incomplete multi-view classification. 
    \item To fully exploit the high-quality imputed data and reduce the affect of the low-quality imputed views, we propose to weight the imputed data from two aspects, avoiding the negative effect on single view and multi-view fusion. The uncertainty-aware training and fusion significantly ensure the effectiveness and reliability of integrating uncertain imputed data.
    \item We conduct experiments on classification for data with multiple types of features or modalities, and evaluate the results with diverse metrics, which validates that the proposed UIMC outperforms existing methods in the above tasks and is trustworthy with reliable uncertainty. %The code\footnote{\url{https://github.com/lblaoke/TLC}} is publicly available.
\end{itemize}

\section{Related Work}
\label{Related Work}
\subsection{Incomplete Multi-View Learning}
Incomplete multi-view learning can be generally divided into two main lines in terms of how to handle missing views. Specifically, existing works mainly focus on neglecting or complementing the missing views based on deep-learning methods.
\textbf{Methods without imputation. }The methods only use present views and directly learn the common latent subspace or representation for all views in clustering~\cite{MCP, PMC, IMVSG} and classification~\cite{CPM, DeepIMV}. 
\textbf{Generative Methods.} The methods impute missing views with the present views and then utilize the reconstructed complete data to conduct downstream tasks~\cite{MVAE, MVTCAE, MIWAE, MCWIV, IMVC, COMPLETER, CPM-GAN, GAN_clu1, GAN_clu2}. Specifically, one of the most popular ways is applying the structure of variational auto-encoder on partial multi-view data to reconstruct missing views~\cite{MVAE, MVTCAE, MIWAE}. Generative adversarial network is also used to generate missing views~\cite{CPM-GAN, GAN_clu1, GAN_clu2}. Besides, there are some methods to obtain imputations based on kernel CCA~\cite{MCWIV}, spectral graph~\cite{IMVC}, and information theory~\cite{COMPLETER}.
Compared with the above algorithms, our method obtains multiple imputations instead of single imputation and then dynamically evaluates the imputation quality. Thus the more reliable downstream classification tasks can be performed.

\subsection{Uncertainty Estimation}
One of the key points of our method is to explore and exploit the uncertainty in missing data. To achieve high-quality uncertainty estimation, many approaches have been proposed \cite{uncertainty_ensemble, uncertainty_prior, uncertainty_deter}. The uncertainty in deep learning can be generally divided into aleatoric uncertainty and epistemic uncertainty~\cite{aleatoric_epistemic1, aleatoric_epistemic2, aleatoric_epistemic3}. Aleatoric uncertainty refers to the uncertainty caused by data and it measures the inherent noise of data. Aleatoric uncertainty can be further divided into homoscedastic uncertainty and heteroscedastic uncertainty, while the first one which varies with different tasks is usually used to estimate the uncertainty in multi-task learning~\cite{multi-task_TUL, multi-task_WL}, and the latter one which varies with input is useful when the input space includes variable noise~\cite{heteroscedastic-deep, heteroscedastic-time}. On the other hand, epistemic uncertainty refers to uncertainty caused by insufficient model training and can be eliminated in theory. It can be estimated by predicting an uncertain observation using models with different parameters, the instability of predicting results just reflects the epistemic uncertainty~\cite{aleatoric_epistemic1, UIDNN}. In this work, we estimate the aleatoric uncertainty of imputations by adopting subjective logic~\cite{LFUP} and Dempster-Shafer theory~\cite{GOBI} to construct a trustworthy and reliable multi-view classification network.

\section{Method}
\label{sec:method}
The key goal of the proposed method is to explore and exploit the uncertainty of incomplete multi-view data, promoting both effectiveness and trustworthiness of the model. We first introduce the background of incomplete multi-view classification in Sec.~\ref{sec:back} and then present how to characterize the uncertainty of imputed view in Sec.~\ref{sec:generation}, exploit the imputation uncertainty and integrate the uncertain decisions in Sec.~\ref{sec:classifier}, and finally demonstrate how to obtain classification predictions with multiple imputations in Sec.~\ref{sec:prediction}.
% (1) a multimodal imputater to impute the missing modalities (in section \ref{sec:generation}); 
% (2) Evidence-guided classifiers linked with Dirichlet distribution and subjective logic for each modality to obtain useful classification predictions (in section \ref{sec:classifier});
% (3) an aleatoric uncertainty-guided weight strategy to force classification network to learn more from high-quality imputed data while less from poor-quality imputed data (in section \ref{sec:integration}); 
% (4) Introducing Dempster's Combination Rule to reasonably fuse different modalities (in section \ref{sec:fusion}).

\subsection{Background}
\label{sec:back}
Given $N$ training inputs $\{\displaystyle \sX_n\}_{n=1}^N$ with $V$ views , i.e., $\displaystyle \sX=\{\mathbf{x}^v\}_{v=1}^{V}$,  and the corresponding class labels $\{\mathbf{y}_{n}\}_{n=1}^N$, multi-view classification aims to construct a mapping between input and label by exploiting the complementary multi-view data. In this paper, we focus on the incomplete multi-view classification task defined in Def.~\ref{def:partial}.
\begin{figure*}
     \centering
     \begin{subfigure}[b]{0.49\textwidth}
         \centering
         \includegraphics[width=\textwidth]{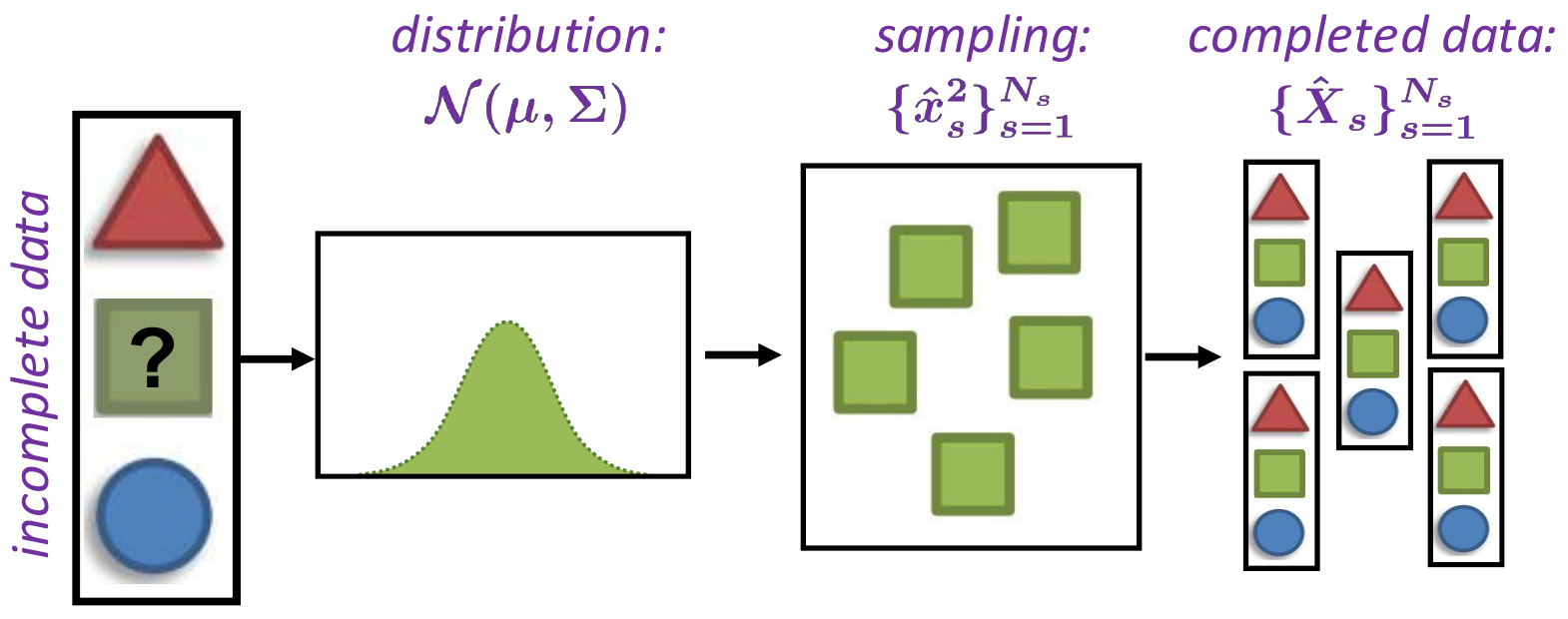}
         \caption{Stage 1: Completion for missing view }
     \label{model1}
     \end{subfigure}
     \hfill
          \begin{subfigure}[b]{0.49\textwidth}
         \centering
         \includegraphics[width=\textwidth]{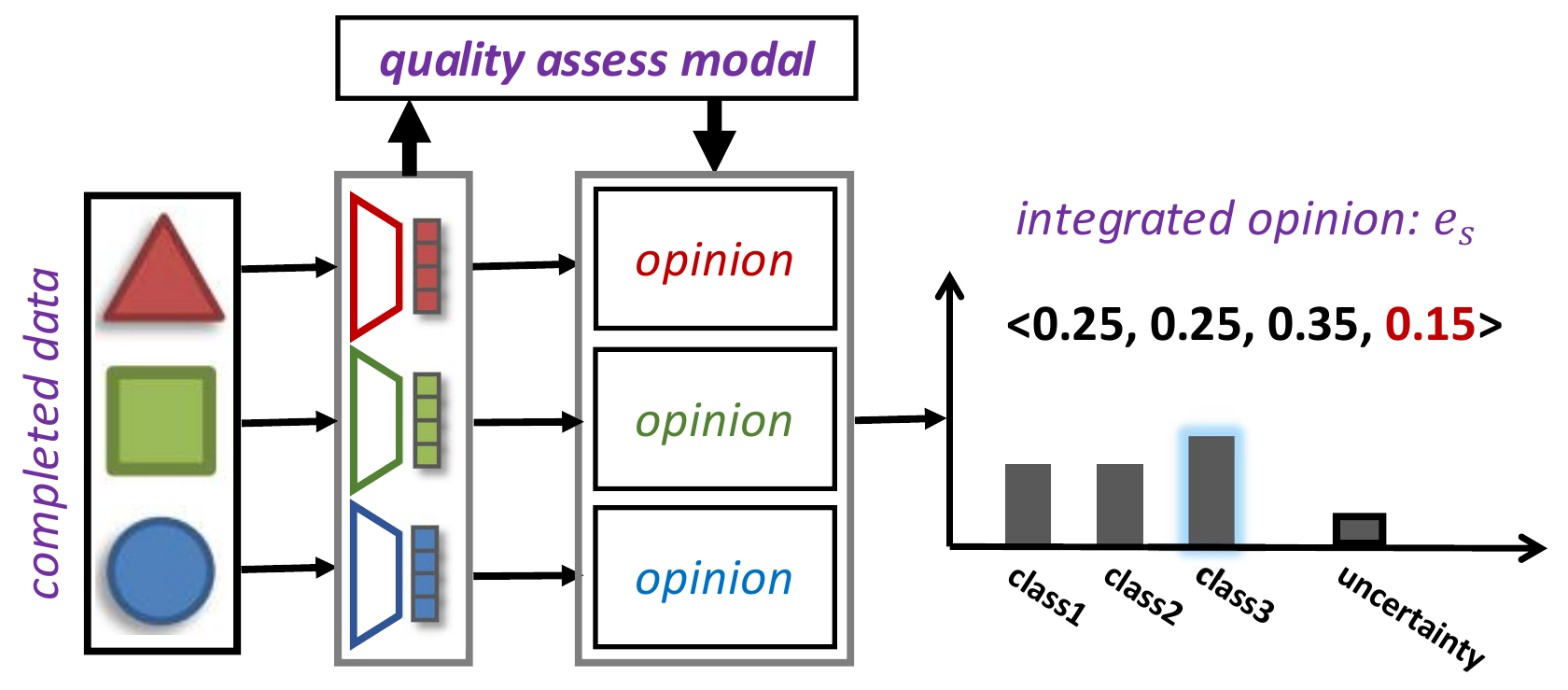}
         \caption{Stage 2: Evidential fusion for completed data}
     \label{model2}
     \end{subfigure}
\caption{Illustration of the proposed UIMC. For clarity, we use 3-view data with one missing view as example in the Fig.~\ref{model1}. The overall process of UIMC consists of two stages. In the stage 1, we characterize the underlying uncertainty of missing view with a  multivariate Gaussian distribution $\mathcal{N}(\boldsymbol{\mu}, \boldsymbol{\Sigma})$ based on the non-parameterized nearest-neighbor strategy. Then we can obtain $N_s$ complete instances $\{\hat{\displaystyle \sX}_{s}\}_{s=1}^{N_s}$ by $N_s$ samplings from the distribution of the missing view.  In the stage 2, for instance $\hat{\displaystyle \sX}_{s}$, to alleviate the negative impact from low-quality imputations, we exploit the uncertainty of each view with evidential classification. Then the Dempster’s combination rule introduced in Def.~\ref{def:dempster_combination} is adopted to integrate the subjective opinions from multiple views, which enables the model to trustworthily exploit the imputation data.}
\label{model}
\end{figure*}

\begin{definition}(Incomplete Multi-View Classification) \label{def:partial}
Formally, a complete multi-view sample is composed of $V$ views $\displaystyle \sX = \{\mathbf{x}^v\}_{v=1}^{V}$ and the corresponding class label $\mathbf{y}$. An incomplete multi-view observation $\overline{\displaystyle \sX}$ is a subset of the complete multi-view observation (i.e., $\overline{\displaystyle \sX} \subseteq \displaystyle \sX$) with arbitrary possible $\overline{V}$ views, where $ 1 \leq \overline{V} \leq V$. Given an incomplete multi-view training dataset $\{\overline{\displaystyle \sX}_n, \mathbf{y}_n\}_{n=1}^{N}$ with $N$ samples, incomplete multi-view classification aims to learn a mapping between the incomplete multi-view observation $\overline{\displaystyle \sX}$ and the corresponding class label $\mathbf{y}$.
\end{definition}

There are two main lines to solve the incomplete multi-view classification problem, including imputing and neglecting the missing views. Specifically, the imputation-based methods~\cite{MVAE,MIWAE,impute_ADMM,impute_adversarial} complete the missing views based on observed data, but they basically ignore that the untrustworthy imputation will harm downstream tasks. In contrast, the methods neglecting the missing views~\cite{DeepIMV,CPM} train the classification model only using the available views, which usually lack of the exploration of the correlation among different views. The proposed method aims to learn a more trustworthy model by exploring and exploiting the uncertainty of the missing views. 
% To provide more trustworthy multi-view classification, we proposed the evidential imputer for partial multimodal classification, which elegantly combines the advantages of the above two schemes through uncertainty estimation. Specifically, the proposed method is composed of 
% (1) a multimodal imputater to impute the missing modalities (in section \ref{sec:generation}); 
% (2) Evidence-guided classifiers linked with Dirichlet distribution and subjective logic for each modality to obtain useful classification predictions (in section \ref{sec:classifier});
% (3) an aleatoric uncertainty-guided weight strategy to force classification network to learn more from high-quality imputed data while less from poor-quality imputed data (in section \ref{sec:integration}); 
% (4) Introducing Dempster's Combination Rule to reasonably fuse different modalities (in section \ref{sec:fusion}).

\subsection{Imputing Missing Views}
\label{sec:generation}
Given an incomplete multi-view data-instance $\overline{\displaystyle \sX}$, we consider reasonably imputing its missing views based on the observed views, which could potentially promote the downstream classification model.

\textbf{Characterizing uncertainty for missing views.} Most of existing incomplete multi-view classification methods~\cite{MVAE,MIWAE,impute_ADMM,impute_adversarial} impute the missing views with single imputation. Formally, they usually construct a deterministic mapping from the incomplete multi-view data-instance $\overline{\displaystyle \sX}$ to complete data, i.e., $f:\overline{\displaystyle \sX}\rightarrow \displaystyle \sX$. The single imputation ignores the high uncertainty nature of the imputed data, which might negatively impact downstream classification tasks due to the untrustworthy imputation. This phenomenon has also been well recognized in single-view classification tasks with missing attributes~\cite{bad_imputation,impute_uncertainty}. Instead of imputing  missing views with deterministic mapping, we characterize the underlying uncertainty of missing views with a distribution based on the nearest-neighbor-based non-parameterized strategy.

\textbf{Constructing the nearest neighbor set.} Without loss of generality, we consider that there is an incomplete multi-view training instance $\overline{\displaystyle \sX}$ and its corresponding label $\mathbf{y}$ with $m$-th view missing, we aim to impute the missing view $\mathbf{x}^m$ based on the information of other training samples whose $m$-th view is available. Inspired by the classical classification algorithm $k$-nearest neighbors~\cite{KNN}, we employ a non-parameterized method to construct the distribution of missing view $\mathbf{x}^m$ by exploring the neighbors of $\overline{\displaystyle \sX}$. Specifically, for the $v$-th view ($v \neq m$), given the available observation $\mathbf{x}^v \in \overline{\displaystyle \sX}$, we construct a neighbor set by finding its $k$-nearest neighbors in other samples with the same label in the following way. Firstly, we construct the distance set $\mathbb{D}^v$ by computing the distance between $\mathbf{x}^v$ and $\mathbf{x}^v_n$, where $\mathbf{x}^v_n$ is the data of $v$-th view of the samples available for $m$-th view and $\mathbf{y}=\mathbf{y}_n$,
\begin{equation}
\mathbb{D}^v=\left\{-\left\|\mathbf{x}^v-\mathbf{x}^v_n\right\|^2 \mid \mathbf{x}^v_n, \mathbf{x}^m_n  \textit{ are available and }  \mathbf{y}=\mathbf{y}_n \right\}.
\end{equation}

Then we can obtain the nearest neighbor indicator set $\mathbb{I}^v$ with
\begin{equation}
\mathbb{I}^v=\left\{i \mid-\left\|\mathbf{x}^v-\mathbf{x}^v_i\right\|^2 \in \operatorname{topk}\left(\mathbb{D}^v\right) \right\},
\end{equation}
where $\operatorname{topk}(\cdot)$ is an operator to select the $k$-nearest neighbors from the training set according to the distance set $\mathbb{D}^v$. Note that, during the test time, the missing view $\mathbf{x}^m$ is imputed by selecting the nearest neighbors from the training set being independent with labels.
% While for test instance $\overline{\displaystyle \sX}$ with missing view $m$, the missing view $x^m$ is imputed by setting $\operatorname{topk}(\cdot)$ as an operator to select the most nearest neighbors from the training set taking no account of label.

\textbf{Imputation with statistical information.} Given the nearest neighbor indicator set $\mathbb{I} = \{\mathbb{I}^v\}_{v\neq m}$, we characterize the distribution of the missing view $\mathbf{x}^m$ with its neighbor set $\{\mathbf{x}_i^m\}_{i\in \mathbb{I}}$ statistically. Formally, we assume that the missing view $\mathbf{x}^m$ follows a multivariate Gaussian distribution $\mathcal{N}(\boldsymbol{\mu}, \boldsymbol{\Sigma})$, where $\boldsymbol{\mu}$ and $\boldsymbol{\Sigma}$ are the mean vector and covariance matrix calculated from neighbor set with
\begin{equation}
\label{eq:statical}
\begin{aligned}
& \boldsymbol{\mu} = \frac{\sum_{i\in \mathbb{I}} \mathbf{x}_i^m}{|\mathbb{I}|}, \boldsymbol{\Sigma} = \frac{1}{|\mathbb{I}|-1} \sum_{i\in\mathbb{I}}{(\mathbf{x}_i^m-\boldsymbol{\mu})(\mathbf{x}_i^m-\boldsymbol{\mu})^T}.
\end{aligned}
\end{equation}
Then $\mathbf{x}^m$ can be imputed by taking multiple samplings from the multivariate Gaussian distribution $\mathcal{N}(\boldsymbol{\mu}, \boldsymbol{\Sigma})$. More specifically, given an incomplete training instance $\overline{\displaystyle \sX}$, we can obtain $N_s$ complete training instances $\{\hat{\displaystyle \sX}_{s}\}_{s=1}^{N_s}$ via $N_s$ samplings. The imputation framework can be referred to Fig.~\ref{model1}.
%Suppose there are $N_1$ complete (i.e., $\bar{V}=V$) observations and $N_2$ incomplete observations in training dataset, size of imputed complete datasets becomes $\hat{N} = N_1+N_2*F$.

\subsection{Classifying Imputed Multi-View Samples}
\label{sec:classifier}
After obtaining imputed multi-view training dataset $\{\{\hat{\displaystyle \sX}_{n,s}\}_{n=1}^{N}\}_{s=1}^{N_s}$, we consider alleviating the negative impact from low-quality imputations by exploiting the uncertainty of imputed samples. Due to inherent noise in imputation, imputed multi-view data are usually of high uncertainty. Therefore we first estimate the uncertainty of different views based on an evidential multi-view learning framework and then conduct uncertainty-based decision fusion. The overall framework is shown in Fig.~\ref{model2}.

Different from traditional classification algorithm, evidential classification \cite{sensoy2018evidential,audun2018subjective} defines a theoretical framework to obtain the subjective opinions $\mathcal{S}=\{\{b_k\}_{k=1}^K, u\}$ including subjective probabilities (belief masses) $\{b_{k}\}_{k=1}^{K} \geq 0 $ and overall uncertainty mass $u\geq 0$, where $K$ denotes the number of classes and $\sum_{k=1}^{K}b_{k}+u = 1$.  The subjective opinion is associated with a Dirichlet distribution with parameters $\boldsymbol{\alpha} = [\alpha_1, \cdots, \alpha_{K}]$. Specifically, a belief mass $b_k$ can be derived easily from the parameters of the corresponding Dirichlet distribution with $b_k = (\alpha_{k}-1)/\alpha_0$, where $\alpha_0=\sum_{k=1}^{K}\alpha_k$ is the Dirichlet strength.

To obtain the Dirichlet distribution parameters for each view ${\hat{\mathbf{x}}}^v \in \hat{\mathbb{X}}$, we construct the evidence neural network by replacing \textit{softmax} layer of traditional neural network classifier with activation function such as \textit{softplus} so that the positive output can be regarded as evidence vector $\boldsymbol{e}^v = [e_1^v, \cdots, e_{K}^v]$. Then the parameters of Dirichlet distribution $Dir(\boldsymbol{p}^v\mid\boldsymbol{\alpha}^v)$ for view ${\hat{\mathbf{x}}}^v$ are obtained with ${\alpha}_k^v = {e}_{k}^v+1$. Under the Variational framework, the evidential classification loss function $\mathcal{L}^v$ is a combination of classification loss $\mathcal{L}^v_c$ and a regularization term $\mathcal{L}^{v}_{r}$ \cite{han2022trusted}. More specifically, the classification objective function can be regarded as an integral of the traditional cross-entropy loss on the simplex determined by $Dir(\boldsymbol{p}^v \mid \boldsymbol{\alpha}^v)$. Specifically,  it is induced as:
\begin{equation}
\begin{aligned}
\mathcal{L}^v_c(\boldsymbol{\alpha}^v\mid\mathbf{x}^v)=\sum_{k=1}^{K} y_{k}\left(\psi\left(\alpha_0^v\right)-\psi\left(\alpha_{k}^{v}\right)\right),
 %= \int\left[\sum_{k=1}^{K}-y_{k} \log \left(\mu_{k}^{m}\right)\right] Dir(\boldsymbol{\mu}^m\mid \boldsymbol{\alpha}^m)d\boldsymbol{\mu}^m 
\end{aligned}
\end{equation}  
where $y_k$ is the $k$-th element of $\mathbf{y}$ represented by a one-hot vector and $\psi(\cdot)$ is the digamma function. To obtain a reasonable Dirichlet distribution, we employ the following equation to add a prior for the Dirichlet distribution as the regularization term: 
\begin{equation}
\begin{aligned}
\mathcal{L}_r^v(\boldsymbol{\alpha}^v\mid\mathbf{x}^v) = D_{KL}\left[Dir(\boldsymbol{p}^v \mid \tilde{\boldsymbol{\alpha}}^v) \|Dir(\boldsymbol{p}^v \mid [1, \cdots, 1]) \right],
% & = \log \left(\frac{\Gamma\left(\sum_{k=1}^{K} \tilde{\alpha}_{k}^m\right)}{\Gamma(K) \prod_{k=1}^{K} \Gamma\left(\tilde{\alpha}_{k}^m\right)}\right)\\
% &\quad+\sum_{k=1}^{K}\left(\tilde{\alpha}_{k}-1\right)\left[\psi\left(\tilde{\alpha}_{k}\right)-\psi\left(\sum_{k=1}^{K} \tilde{\alpha}_k^m\right)\right],
\end{aligned}
\end{equation}
where $\tilde{\boldsymbol{\alpha}}^v = \mathbf{y}+\left(1-\mathbf{y}\right) \odot \boldsymbol{\alpha}^v$ is the Dirichlet distribution after replacing the $\alpha_k$ corresponding to the label with $1$. $Dir(\boldsymbol{p}^v \mid [1, \cdots, 1])$ is the uniform Dirichlet distribution.
Then for each view the overall loss function can be written as 
\begin{equation}
\mathcal{L}^v(\boldsymbol{\alpha}^v\mid\mathbf{x}^v) = \mathcal{L}^v_{c}(\boldsymbol{\alpha}^v\mid\mathbf{x}^v) + \lambda \mathcal{L}_{r}^v(\boldsymbol{\alpha}^v\mid\mathbf{x}^v),
\end{equation}
where $\lambda$ is the annealing coefficient gradually changing from 0 to 1 during training to control constraint strength. 
We could obtain the Dirichlet distribution $Dir(\boldsymbol{p}^v\mid \boldsymbol{\alpha}^v)$ by minimizing $\mathcal{L}^v$. Then the corresponding subjective opinions could be derived with $b_k^v = ({\alpha}_{k}^v-1)/\alpha_0$ and $u^v=1-\sum_{k=1}^{K}b_k^v$.

Now we consider integrating the subjective opinions from different views according to the uncertainty. Dempster-Shafer theory allows to integrate subjective opinions from different sources to produce a more comprehensive opinion \cite{dempster1968generalization}. Specifically, the Dempster's combination rule for two different views is defined in Def.~\ref{def:dempster_combination}. Accordingly, the integrated multi-view subjective opinion can be derived with $\mathcal{S}^m = \mathcal{S}^1 \oplus \mathcal{S}^2\oplus \cdots \mathcal{S}^V = \{\{b_k^m\}_{k=1}^K, {u}^m\}$. The corresponding parameters of the integrated multi-view Dirichlet distribution $Dir(\boldsymbol{p}^m\mid \boldsymbol{\alpha}^m)$ are obtained with ${\alpha}_k^m = {b}^m_{k} \times \alpha_0^m+1$.

\begin{definition} 
\label{def:dempster_combination}
(Dempster's Combination Rule.) The subjective opinion $\mathcal{S}=\{\{b_k\}_{k=1}^K, u\}$ after integration can be obtained from two subjective opinion $\mathcal{S}^1=\{\{b^1_k\}_{k=1}^K, u^1\}$ and $\mathcal{S}^2=\{\{b^2_k\}_{k=1}^K, u^2\}$ with $\mathcal{S}=\mathcal{S}^1\oplus\mathcal{S}^2$. Specifically, it can be written as:
% \begin{equation}
% \mathcal{S} = \mathcal{S}^1 \oplus \mathcal{S}^2, \
% \end{equation}
% where
\begin{equation}
b_{k}=\frac{1}{1-C}\left(b_{k}^{1} b_{k}^{2}+b_{k}^{1} u^{2}+b_{k}^{2} u^{1}\right), u=\frac{1}{1-C} u^{1} u^{2},
\end{equation}
where $C = \sum_{i \neq j}b_i^1b_j^2$. 
\end{definition}
To obtain reliable both single-view and integrated Dirichlet distribution, we employ a multi-task strategy. Specifically,  the final loss function in classification period composes of which of single and multiple views:
\begin{equation}
\begin{aligned}
\label{eq:loss}
&\mathcal{L} = \sum_{n=1}^N\sum_{s=1}^{N_s} \left\{\mathcal{L}^{m}(\boldsymbol{\alpha}^m)+\sum_{v=1}^{V}\mathcal{L}^{v}(\boldsymbol{\alpha}^v\mid\mathbf{x}^v_{n,s})\right\},\\
\end{aligned}
\end{equation}
where
\begin{equation}
\begin{aligned}
\mathcal{L}^{m}&(\boldsymbol{\alpha}^m) = \sum_{k=1}^{K} y_{k}\left(\psi\left(\alpha^m_0\right)-\psi\left(\alpha^m_{k}\right)\right)\\
& + \lambda D_{KL}\left[Dir(\boldsymbol{p}^m \mid \tilde{\boldsymbol{\alpha}}^m) \|Dir(\boldsymbol{p}^m \mid [1, \cdots, 1]) \right].
\end{aligned}
\end{equation}
The pseudo-code of UIMC is provided in Alg.~\ref{alg:code_train}.

\begin{algorithm}[!htbp]
\caption{Training Process of UIMC}
\label{alg:code_train}
\SetAlgoLined 
%\textbf{/*Training*/}\\
\textbf{Input: }Incomplete multi-view training set $\{\overline{\displaystyle \sX}_n, \mathbf{y}_n\}_{n=1}^N$.\\
\textbf{Initialize:}
\text{Initialize the parameters of UIMC.}\\
\For {$n=1:N$}{
        Impute missing view $x_n^m$ by taking $N_s$ samplings from distribution $\mathcal{N}(\boldsymbol{\mu}, \boldsymbol{\Sigma})$ with Eq.~\ref{eq:statical}.\\
    }
\While{not converged}{
        \For {$v=1:V$}{
            Obtain evidence ${\boldsymbol{e}}_n^v$ with DNN;\\
            Obtain $\boldsymbol{\alpha}_n^v = {\boldsymbol{e}}_n^v + 1$ and $\mathcal{S}_n^v=\{\{b_{n,k}^v\}_{k=1}^K, u_n^v\}$.\\
        }
        Obtain ${\boldsymbol{\alpha}}_{n} = [\alpha_{n,1}, \cdots, \alpha_{n,K}]$ and $\mathcal{S}_n = \{\{b_{n,k}\}_{k=1}^K, u_n\}$ through Def.~\ref{def:dempster_combination}.\\
    Calculate overall loss $\mathcal{L}$ in Eq. \ref{eq:loss};\\
    Use an optimized algorithm to update the parameters.\\
}
\textbf{Output:}
\text{Parameters of model.}\\
\end{algorithm}
\subsection{{Predicting with Multiple Imputations}}
\label{sec:prediction}
During the test stage, for each incomplete sample, we could obtain $N_s$ imputed multi-view samples $\{\hat{\displaystyle \sX}_{s}\}_{s=1}^{N_s}$. Then we employ a voting strategy to acquire the classification prediction with $N_s$ imputations for each sample. Specifically, given the $N_s$ imputed test instances $\{\hat{\displaystyle \sX}_{s}\}_{s=1}^{N_s}$, we can obtain $N_s$ label predictions $\{\hat{y}_s\}_{s=1}^{N_s}$. The final predict label for multi-view test instance $\hat{\displaystyle \sX}$ would be the most frequent element in $\{\hat{y}_s\}_{s=1}^{N_s}$.
\section{Experiments}
\label{Experiments}
In this section, we conduct extensive experiments on multiple multi-view datasets with various missing rates $\eta = \frac{\sum_{v=1}^{V}{M}^v}{V \times N}$ to answer the following questions, where $M^v$ indicates the number of observations without the $v$-th view. 
Q1 Effectiveness (I). Is the proposed UIMC superior to other methods?
Q2 Effectiveness (II). How about the quality of the imputed complete multi-view data of different methods?
Q3 Ablation study (I). Is multiple-imputation strategy really better than single-imputation strategy?
Q4 Ablation study (II). Do the uncertainty-guided classification and fusion strategy help with performance?
Q5 Stability. How about the stability of the proposed method under the sampling operation?
%Q6 Flexibility. Whether UIMC could be suitable for arbitrary missing patterns? 

\subsection{Datasets and Comparison Methods}
\label{dataset}
To validate the effectiveness of UIMC, we conduct experiments on five datasets.
\textbf{YaleB}~\cite{YaleB} is a 3-view dataset contains 10 categories, and there are 65 facial images in each category. \textbf{ROSMAP}~\cite{ROSMAP} is a 3-view dataset contains two categories: Alzheimer's disease (AD) patients with 182 samples and normal control (NC) with 169 samples. \textbf{Handwritten}~\cite{Handwritten} is a 6-view dataset contains 10 categories from digit ``0'' to ``9''. The number of samples in each category is 200. \textbf{BRCA}~\cite{ROSMAP} is a 3-view dataset for Breast Invasive Carcinoma (BRCA) subtype classification, and it contains 5 categories. The number of samples in each category is between 46 and 436. \textbf{Scene15}~\cite{Scene} is a 3-view dataset contains 15 categories for scene classification, and the number of samples in each category is between 210 and 410.

We compare the proposed UIMC with the following methods:
(1) \textbf{Mean-Imputation} simply imputes missing view $\mathbf{x}^m$ with the mean of all available observations on the $m$-th view. 
(2) \textbf{GCCA}~\cite{GCCA} extends Canonical Correlation Analysis (CCA)~\cite{CCA} to handle data with more than two views. 
(3) \textbf{TCCA}~\cite{TCCA} obtains a common subspace shared by all views by maximizing the canonical correlation of multiple views.
(4) \textbf{MVAE}~\cite{MVAE} extends variational autoencoder to multi-view data and adopts product-of-experts strategy to obtain a common latent subspace. 
(5) \textbf{MIWAE}~\cite{MIWAE} extends importance-weighted autoencoder to multi-view data to impute missing data.
(6) \textbf{CPM-Nets}~\cite{CPM} directly learns the joint latent representations for all views with available data, and maps the latent representation to classification predictions.
(7) \textbf{DeepIMV}~\cite{DeepIMV} applies the information bottleneck (IB) framework to obtain marginal and joint representations with the available data, and constructs the view-specific and multi-view predictors to obtain the classification predictions.

\subsection{Experiment Results}
\label{experimet}
We now provide detailed empirical results to investigate the above key questions, which can validate the effectiveness and trustworthiness of our model.
\begin{table*}[htbp]
\centering
\caption{Comparison in terms of classification accuracy (mean$\pm$std) with $\eta = [0, 0.1, 0.2, 0.3, 0.4, 0.5]$ on five datasets.} % 添加标题
\label{tab:acc} % 设置标签
\resizebox{1.0\textwidth}{!}{
\begin{tabular}{c|c|ccccccc}
\toprule
\multirow{2}{*}{Datasets}    & \multirow{2}{*}{Methods} & \multicolumn{6}{c}{Missing rates}                                                                         \\
% \cline{3-8}
                             &                          & $\eta= 0$               & $\eta=0.1$             & $\eta=0.2$             &$\eta=0.3$             & $\eta=0.4$             & $\eta=0.5$             \\
\midrule
\multirow{8}{*}{YaleB}       & Mean-Imputation                    & $\pmb{{1.0000\pm0.00}}$ & $0.9923\pm0.00$          & $0.9769\pm0.01$          & $0.9769\pm0.01$          & $0.9692\pm0.01$          & $0.9615\pm0.01$          \\
                             & GCCA                     & $0.9692\pm0.00$          & $0.9385\pm0.01$          & $0.9077\pm0.02$          & $0.8615\pm0.03$          & $0.8385\pm0.02$          & $0.8231\pm0.02$          \\
                             & TCCA                     & $0.9846\pm0.00$          & $0.9625\pm0.00$          & $0.9492\pm0.01$          & $0.90.77\pm0.01$          & $0.8846\pm0.02$          & $0.8615\pm0.01$          \\
                             & MVAE                     & $\pmb{1.0000\pm0.00}$ & $0.9969\pm0.00$          & $0.9861\pm0.00$          & $0.9831\pm0.01$          & $0.9692\pm0.02$          & $0.9599\pm0.01$          \\
                             & MIWAE                    & $\pmb{1.0000\pm0.00}$ & $0.9923\pm0.00$          & $0.9923\pm0.01$          & $0.9903\pm0.01$          & $0.9846\pm0.01$          & $0.9692\pm0.03$          \\
                             & CPM-Nets                      & $0.9915\pm0.02$          & $0.9862\pm0.01$          & $0.9800\pm0.01$          & $0.9700\pm0.02$          & $0.9469\pm0.01$          & $0.9100\pm0.02$          \\
                             & DeepIMV                  & $\pmb{1.0000\pm0.00}$ & $0.9846\pm0.03$          & $0.9231\pm0.02$          & $0.9154\pm0.08$          & $0.8923\pm0.02$          & $0.8718\pm0.06$          \\ \cmidrule{2-8}
                             & Ours                      & $\pmb{1.0000\pm0.00}$ & $\pmb{1.0000\pm0.00}$ & $\pmb{0.9981\pm0.00}$ & $\pmb{0.9962\pm0.01}$ & $\pmb{0.9847\pm0.01}$ & $\pmb{0.9769\pm0.01}$ \\
\midrule
\multirow{8}{*}{ROSMAP}      & Mean-Imputation                    & $0.7429\pm0.03$          & $0.6809\pm0.01$          & $0.6714\pm0.02$          & $0.6571\pm0.07$          & $0.6429\pm0.02$          & $0.6072\pm0.05$          \\
                             & GCCA                     & $0.6953\pm0.03$          & $0.6571\pm0.02$          & $0.6429\pm0.04$          & $0.6143\pm0.03$          & $0.5714\pm0.06$          & $0.5429\pm0.06$          \\
                             & TCCA                     & $0.7143\pm0.03$          & $0.7072\pm0.01$          & $0.6857\pm0.05$          & $0.6500\pm0.05$          & $0.6286\pm0.03$          & $0.6036\pm0.06$          \\
                             & MVAE                     & $0.7429\pm0.02$          & $0.7286\pm0.05$          & $0.7143\pm0.05$          & $0.6786\pm0.03$          & $0.6786\pm0.06$          & $0.6524\pm0.06$          \\
                             & MIWAE                    & $0.7429\pm0.03$          & $0.7286\pm0.02$          & $0.6714\pm0.05$          & $0.6571\pm0.03$          & $0.6571\pm0.05$          & $0.6357\pm0.08$          \\
                             & CPM-Nets                      & $0.7840\pm0.05$          & $0.7517\pm0.04$          & $0.7394\pm0.06$          & $0.7183\pm0.04$          & $0.6901\pm0.08$          & $0.6409\pm0.08$          \\
                             & DeepIMV                  & $0.7607\pm0.03$          & $0.7429\pm0.01$          & $0.7143\pm0.05$          & $0.6643\pm0.05$          & $0.6524\pm0.06$          & $0.6250\pm0.06$          \\\cmidrule{2-8}
                             & Ours                      & $\pmb{0.8714\pm0.00}$ & $\pmb{0.8429\pm0.03}$ & $\pmb{0.7714\pm0.05}$ & $\pmb{0.7464\pm0.03}$ & $\pmb{0.7214\pm0.03}$ & $\pmb{0.7143\pm0.02}$ \\
\midrule
\multirow{8}{*}{Handwritten} & Mean-Imputation                    & $0.9800\pm0.00$          & $0.9750\pm0.00$          & $0.9700\pm0.01$          & $0.9700\pm0.01$          & $0.9500\pm0.01$          & $0.9100\pm0.01$          \\
                             & GCCA                     & $0.9500\pm0.01$          & $0.9350\pm0.02$          & $0.9100\pm0.01$          & $0.8875\pm0.02$          & $0.8425\pm0.02$          & $0.8200\pm0.03$          \\
                             & TCCA                     & $0.9725\pm0.00$          & $0.9650\pm0.00$          & $0.9575\pm0.02$          & $0.9350\pm0.01$          & $0.9200\pm0.01$          & $0.9100\pm0.02$          \\
                             & MVAE                     & $0.9800\pm0.00$          & $0.9750\pm0.01$          & $0.9700\pm0.00$          & $0.9650\pm0.01$          & $0.9575\pm0.01$          & $0.9500\pm0.01$          \\
                             & MIWAE                    & $0.9800\pm0.00$          & $0.9800\pm0.00$         & $0.9725\pm0.00$          & $0.9650\pm0.00$          & $0.9475\pm0.01$          & $0.9375\pm0.02$         \\
                             & CPM-Nets                      & $0.9550\pm0.01$          & $0.9475\pm0.01$          & $0.9375\pm0.01$          & $0.9300\pm0.02$          & $0.9225\pm0.01$          & $0.9125\pm0.01$          \\
                             & DeepIMV                  & $\pmb{0.9908\pm0.04}$ & $\pmb{0.9883\pm0.02}$ & $\pmb{0.9850\pm0.04}$ & $0.9750\pm0.02$ & $0.9625\pm0.04$          & $0.9450\pm0.06$          \\ \cmidrule{2-8}
                             & Ours                      & $0.9825\pm0.00$          & $0.9800\pm0.00$         & $0.9800\pm0.00$          & $\pmb{0.9775\pm0.00}$ & $\pmb{0.9700\pm0.01}$ & $\pmb{0.9600\pm0.01}$ \\
\midrule
\multirow{8}{*}{BRCA}        & Mean-Imputation                    & $0.7885\pm0.02$          & $0.7143\pm0.03$          & $0.7000\pm0.04$          & $0.6571\pm0.02$          & $0.6429\pm0.03$          & $0.6286\pm0.02$          \\
                             & GCCA                     & $0.7371\pm0.03$          & $0.7143\pm0.03$          & $0.6971\pm0.04$          & $0.6762\pm0.02$          & $0.6514\pm0.03$          & $0.6381\pm0.04$          \\
                             & TCCA                     & $0.7543\pm0.02$          & $0.7314\pm0.03$          & $0.7238\pm0.04$          & $0.7129\pm0.03$          & $0.6857\pm0.04$          & $0.6743\pm0.03$          \\
                             & MVAE                     & $0.7885\pm0.03$ &$0.7691\pm0.02$ & $0.7347\pm0.01$          & $0.6968\pm0.03$          & $0.6633\pm0.05$          & $0.6388\pm0.03$          \\
                             & MIWAE                    & $0.7885\pm0.02$          & $0.7352\pm0.03$          & $0.7314\pm0.03$          & $0.7105\pm0.02$          & $0.7029\pm0.02$          & $0.6857\pm0.04$          \\
                             & CPM-Nets                      & $0.7388\pm0.02$          & $0.7317\pm0.04$          & $0.7107\pm0.08$          & $0.7233\pm0.04$          & $0.6980\pm0.05$          & $0.6788\pm0.03$          \\
                             & DeepIMV                  & $0.7686\pm0.03$          & $0.7614\pm0.02$          & $0.7457\pm0.02$          & $0.7414\pm0.02$          & $0.7400\pm0.02$          & $0.6714\pm0.04$          \\\cmidrule{2-8}
                             & Ours                      & $\pmb{0.8286\pm0.01}$ & $\pmb{0.7943\pm0.01}$          & $\pmb{0.7771\pm0.01}$ & $\pmb{0.7657\pm0.02}$ & $\pmb{0.7543\pm0.02}$ & $\pmb{0.7429\pm0.02}$ \\
\midrule
\multirow{8}{*}{Scene15}       & Mean-Imputation                    & $0.7681\pm0.02$          & $0.6912\pm0.01$          & $0.6477\pm0.01$          & $0.6098\pm0.01$          & $0.5864\pm0.02$          & $0.5106\pm0.02$          \\
                             & GCCA                     & $0.6611\pm0.02$          & $0.6511\pm0.01$          & $0.6176\pm0.01$          & $0.5708\pm0.01$          & $0.5385\pm0.02$          & $0.5006\pm0.02$          \\
                             & TCCA                     & $0.6878\pm0.02$          & $0.6644\pm0.01$          & $0.6566\pm0.01$          & $0.6187\pm0.01$          & $0.5741\pm0.01$          & $0.5563\pm0.02$          \\
                             & MVAE                     & $0.7681\pm0.00$ & $0.7346\pm0.01$ & $0.7157\pm0.01$ & $0.6689\pm0.01$          & $0.6444\pm0.01$          & $0.6098\pm0.01$          \\
                             & MIWAE                    & $0.7681\pm0.03$          & $0.7179\pm0.01$          & $0.6990\pm0.01$         & $0.6566\pm0.01$          & $0.6265\pm0.02$          & $0.5875\pm0.02$          \\
                             & CPM-Nets                      & $0.6990\pm0.02$          & $0.6566\pm0.02$          & $0.6388\pm0.00$          & $0.6265\pm0.01$          & $0.5903\pm0.01$          & $0.5708\pm0.01$          \\
                             & DeepIMV                  & $0.7124\pm0.00$          & $0.6934\pm0.02$          & $0.6656\pm0.01$          & $0.6410\pm0.00$          & $0.5853\pm0.02$          & $0.5719\pm0.01$          \\\cmidrule{2-8}
                             & Ours                      & $\pmb{0.7770\pm0.00}$ & $\pmb{0.7581\pm0.01}$ & $\pmb{0.7347\pm0.00}$ & $\pmb{0.6990\pm0.01}$ & $\pmb{0.6689\pm0.01}$ & $\pmb{0.6254\pm0.02}$\\
\bottomrule
\end{tabular}}
\end{table*}

\begin{figure*}
     \centering
     \begin{subfigure}[b]{0.32\textwidth}
         \centering
         \includegraphics[width=\textwidth,height=0.48\textwidth]{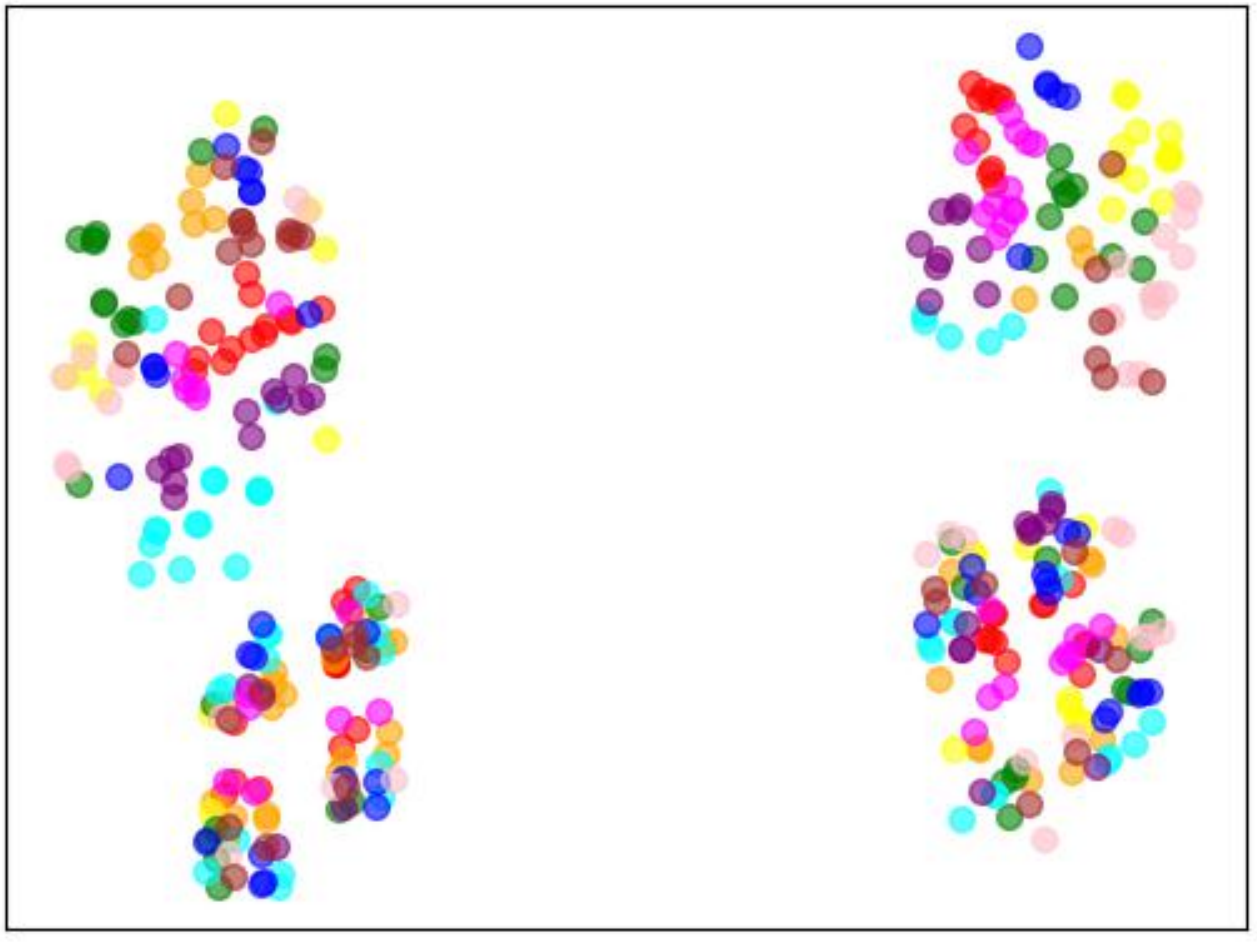}
         \caption{{Impute with zero (O)}}
     \end{subfigure}
     \hfill
     \begin{subfigure}[b]{0.32\textwidth}
         \centering
         \includegraphics[width=\textwidth,height=0.48\textwidth]{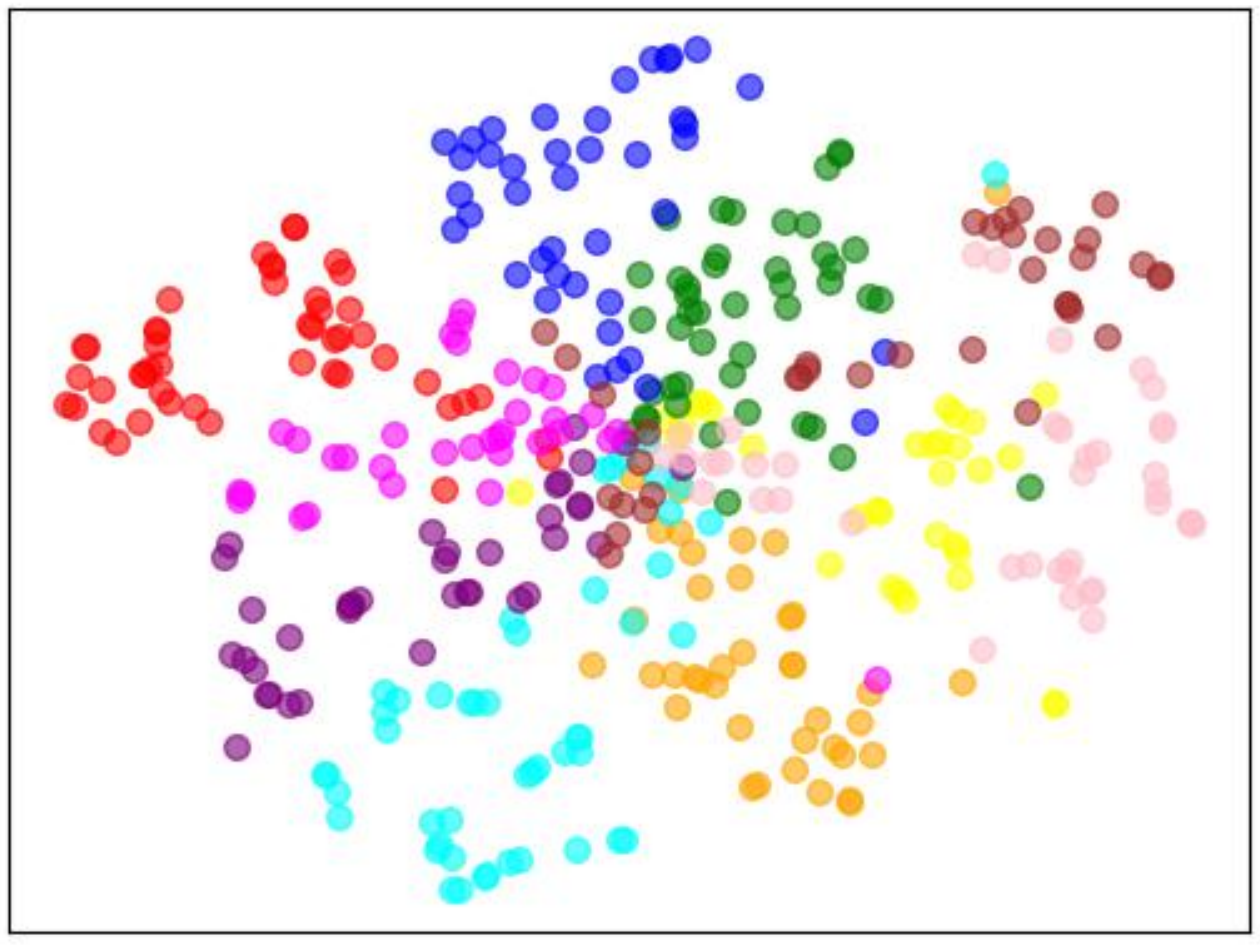}
         \caption{{Impute with mean (O)}}
     \end{subfigure}
     \hfill
     \begin{subfigure}[b]{0.32\textwidth}
         \centering
         \includegraphics[width=\textwidth,height=0.48\textwidth]{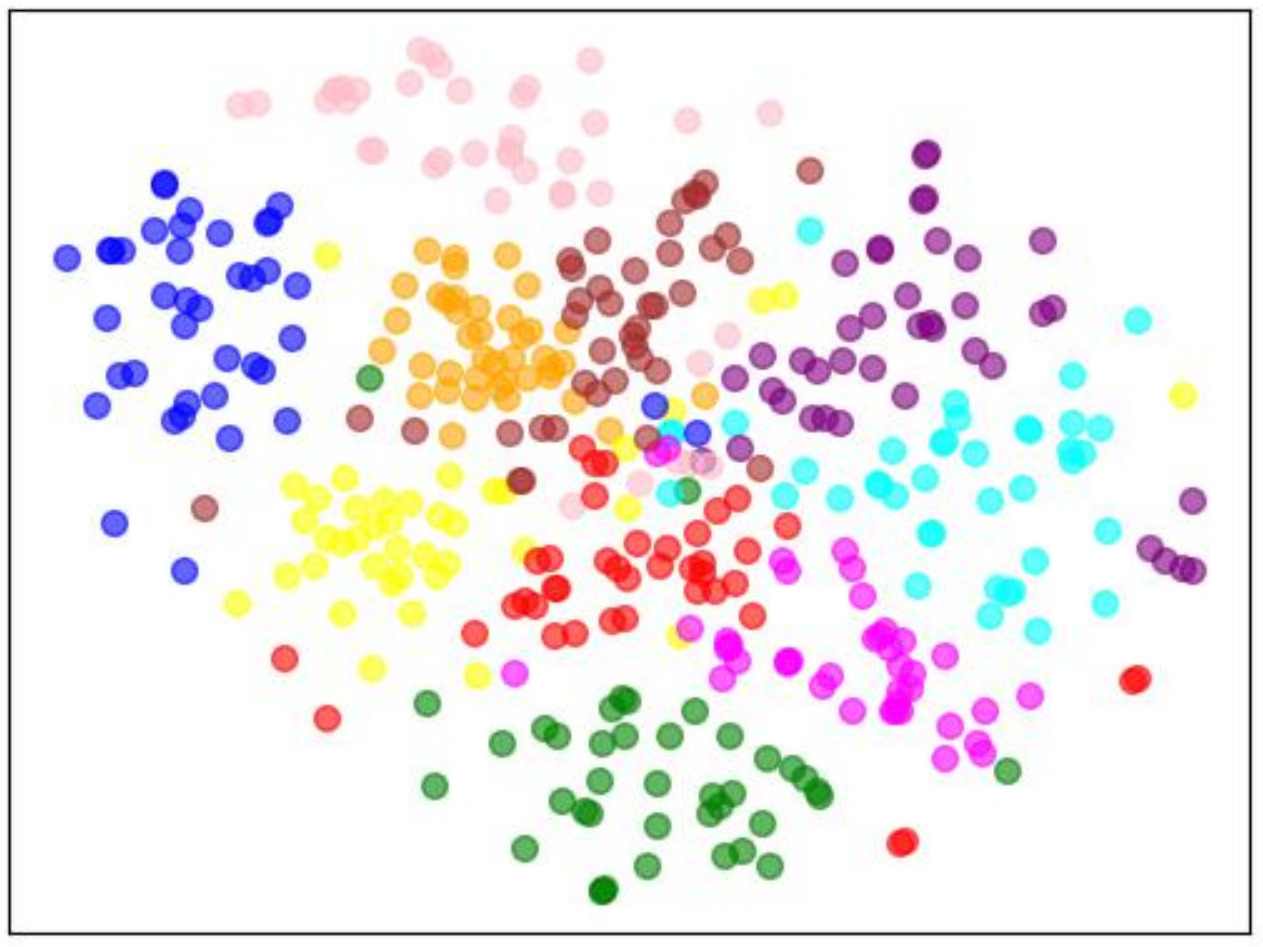}
         \caption{{CPM-Nets (Z)}}
     \end{subfigure}
     \hfill
     \begin{subfigure}[b]{0.32\textwidth}
         \centering
         \includegraphics[width=\textwidth,height=0.48\textwidth]{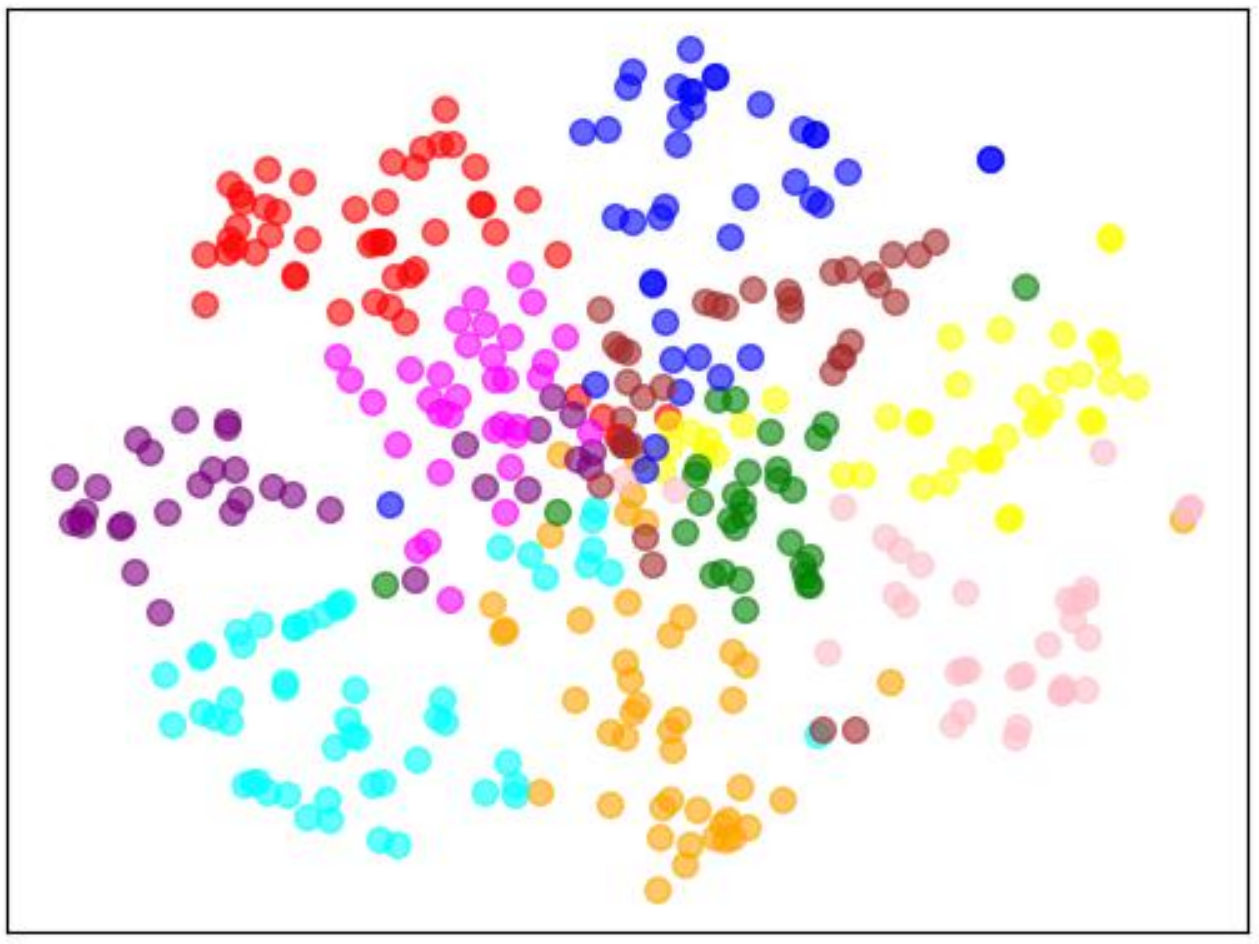}
         \caption{{MVAE (O)}}
     \end{subfigure}
     \hfill
     \begin{subfigure}[b]{0.32\textwidth}
         \centering
         \includegraphics[width=\textwidth,height=0.48\textwidth]{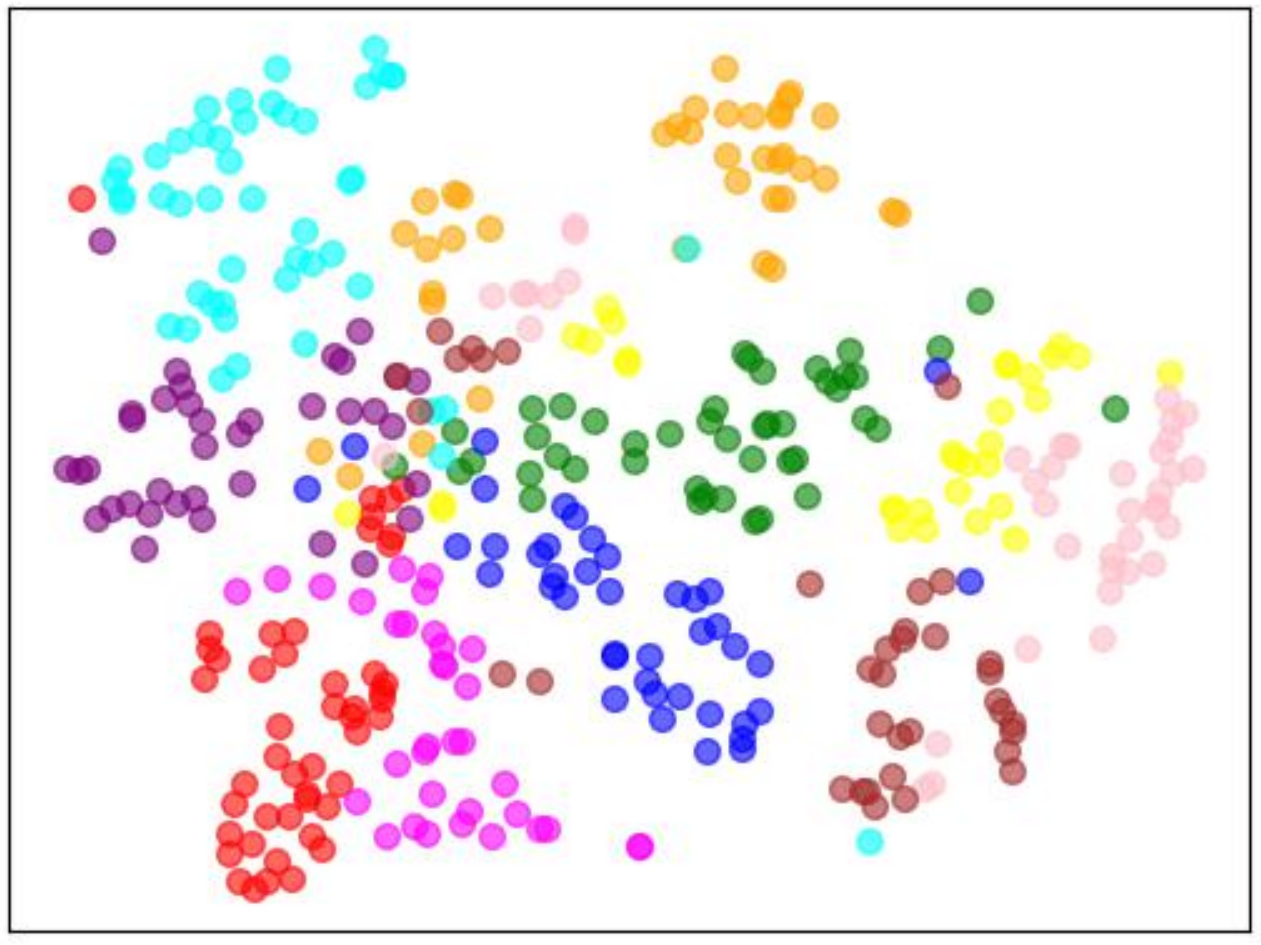}
         \caption{{MIWAE (O)}}
     \end{subfigure}
     \hfill
          \begin{subfigure}[b]{0.32\textwidth}
         \centering
         \includegraphics[width=\textwidth,height=0.48\textwidth]{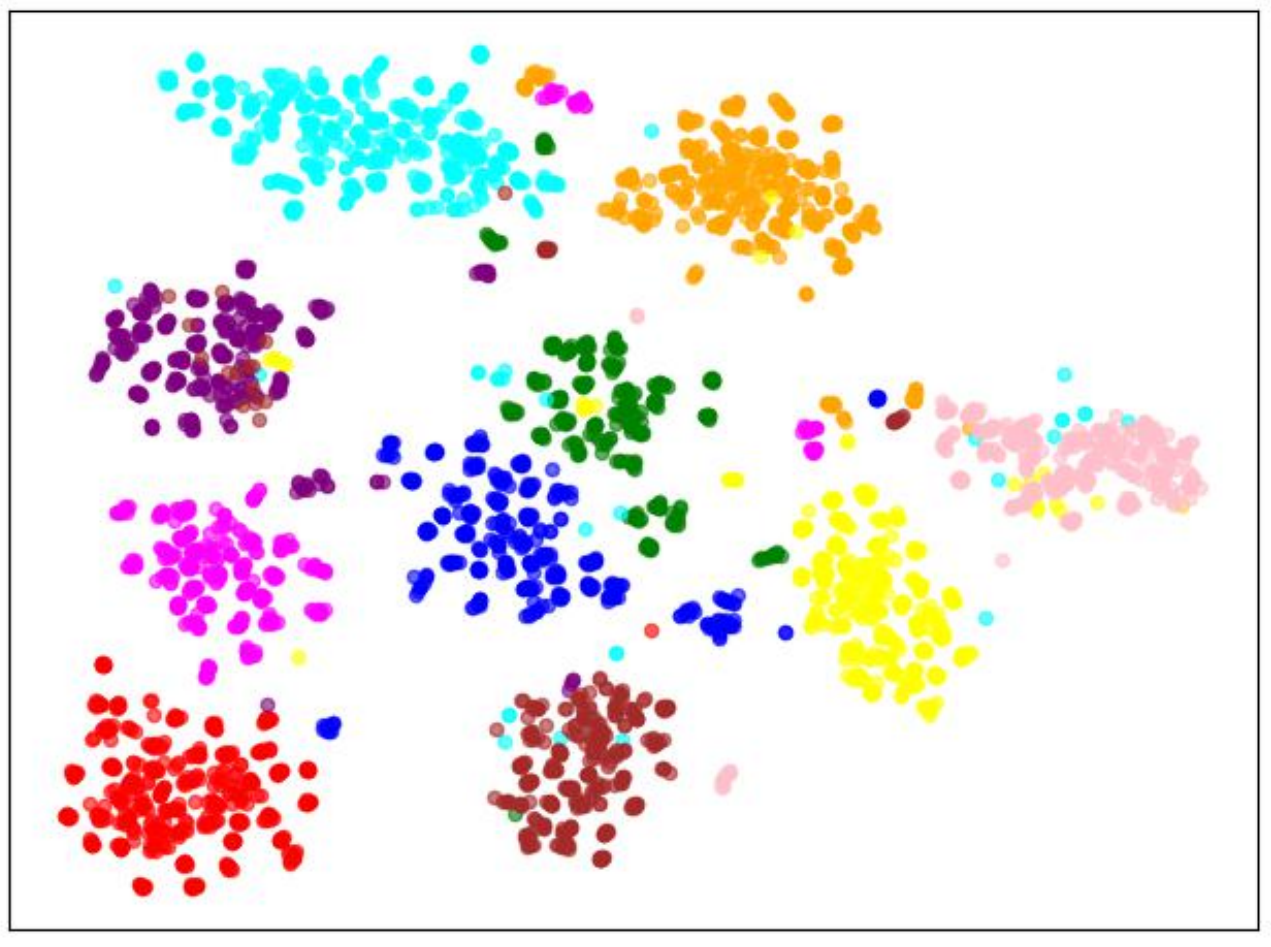}
         \caption{{Ours (O)}}
     \end{subfigure}
        \caption{Visualization of imputed complete data (Handwritten) under $\eta = 0.5$, where multiple views are concatenated. ``O'' and ``Z'' indicate concatenation with original data (and imputed views) and learned common latent representation from multiple available views, respectively.}
        \label{cluster}
\end{figure*}
\textbf{Q1 Effectiveness (I).} We evaluate our algorithm by comparing it with the state-of-the-art incomplete multi-view classification methods with different missing rates $\eta = [0, 0.1, 0.2, 0.3, 0.4, 0.5]$. We employ classification accuracy as the evaluation metric following prior works~\cite{accuracy}. 
For the Mean-Imputation method, we impute the missing views with corresponding means and then train classifiers based on the completed multi-view data. For multi-view learning methods including GCCA and TCCA, we first impute the missing views with the corresponding means, and then train GCCA/TCCA to obtain the joint concatenated representation for all views. Finally, we train a classifier based on the multi-view representation. For incomplete multi-view generative methods including MVAE and MIWAE, we take an impute-then-classify strategy. Specifically, we train MVAE/MIWAE to obtain the imputed multi-view data, and then train classifiers both on single-view and multi-view data based on the imputation. 
For the classification methods that can directly work on incomplete multi-view data including CPM-Nets and DeepIMV, we train the classifiers on the incomplete data without imputing missing views. We adopt the same network architecture for all methods, and the comparison results are shown in Table~\ref{tab:acc}. From the empirical results, it is observed: (1) UIMC achieves competitive performance on complete multi-view data ($\eta = 0$). (2) UIMC outperforms all comparison methods on most datasets when $\eta \neq 0$ especially when the task is difficult. For example, the performance of the proposed method on ROSMAP from $\eta = 0.1$ to $\eta = 0.5$ is significantly superior to other methods. (3) UIMC achieves the highest classification accuracy with relatively high missing rates which validates that UIMC is very robust to incomplete multi-view data. For example, the accuracy declines only $2.28\%$ on BRCA from $\eta = 0.3$ to $\eta = 0.5$ compared with $7\%$ of other methods.

\textbf{Q2 Effectiveness (II).} We conduct experiments for visualizing the imputed complete multi-view data on Handwritten with $\eta = 0.5$. Specifically, we take $N_s=30$ samplings from the distribution according to the available views to obtain the imputed complete data. As shown in Fig.~\ref{cluster}, the margins between different classes are much clearer since UIMC explores the correlation among different views. Although our imputation strategy may introduce noise, our model still achieves the best performance. 

\begin{figure*}[htb]
     \centering
     \begin{subfigure}[b]{0.32\textwidth}
         \centering
         \includegraphics[width=\textwidth]{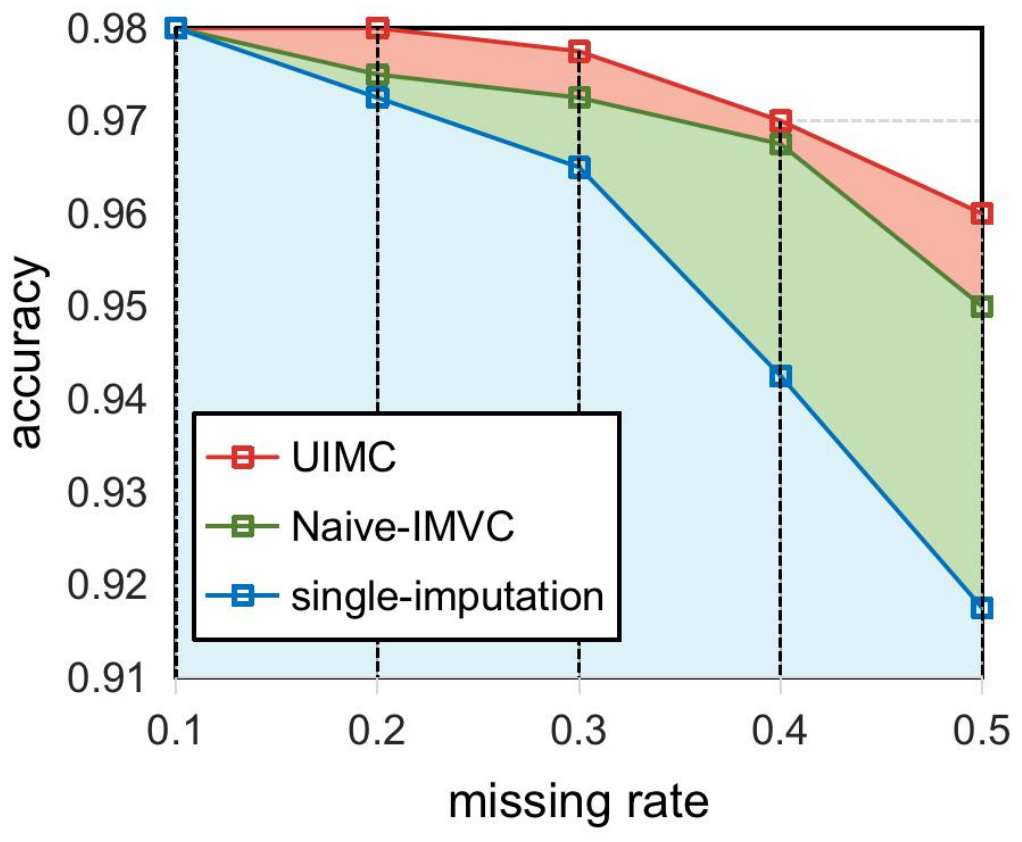}
         \caption{Handwritten}
         \label{fig:y equals x}
     \end{subfigure}
     \hfill
     \begin{subfigure}[b]{0.32\textwidth}
         \centering
         \includegraphics[width=\textwidth]{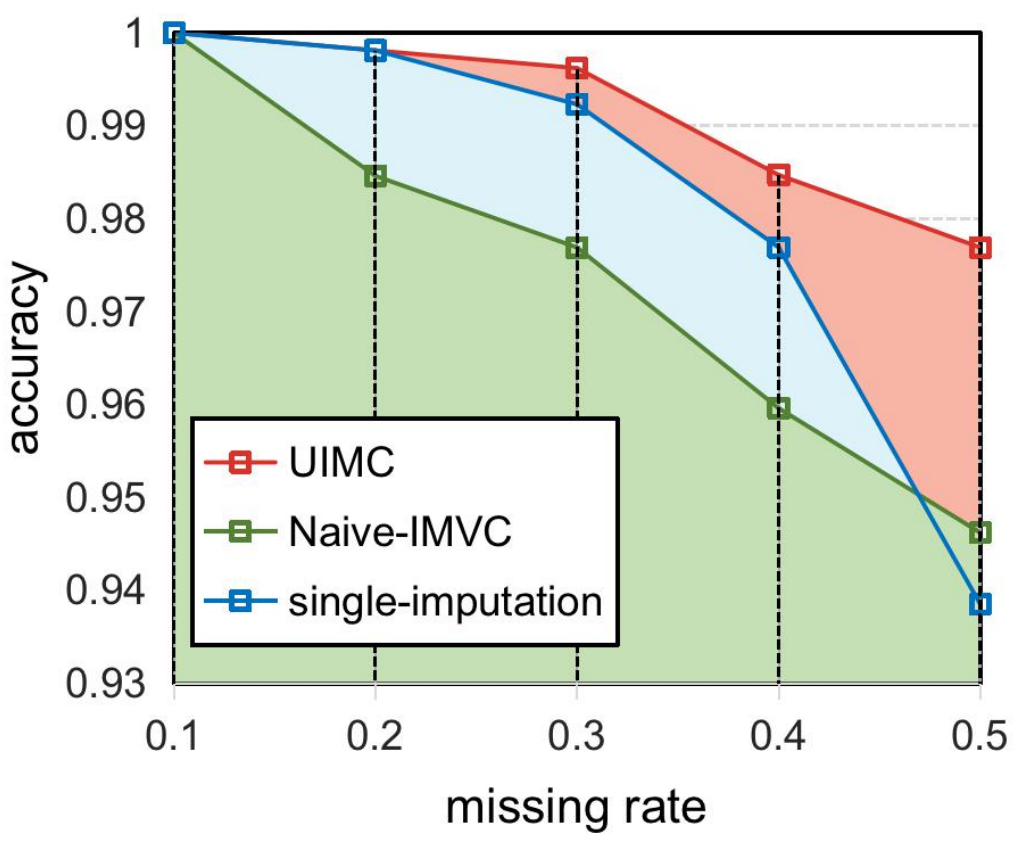}
         \caption{YaleB}
     \end{subfigure}
     \hfill
     \begin{subfigure}[b]{0.32\textwidth}
         \centering
         \includegraphics[width=\textwidth]{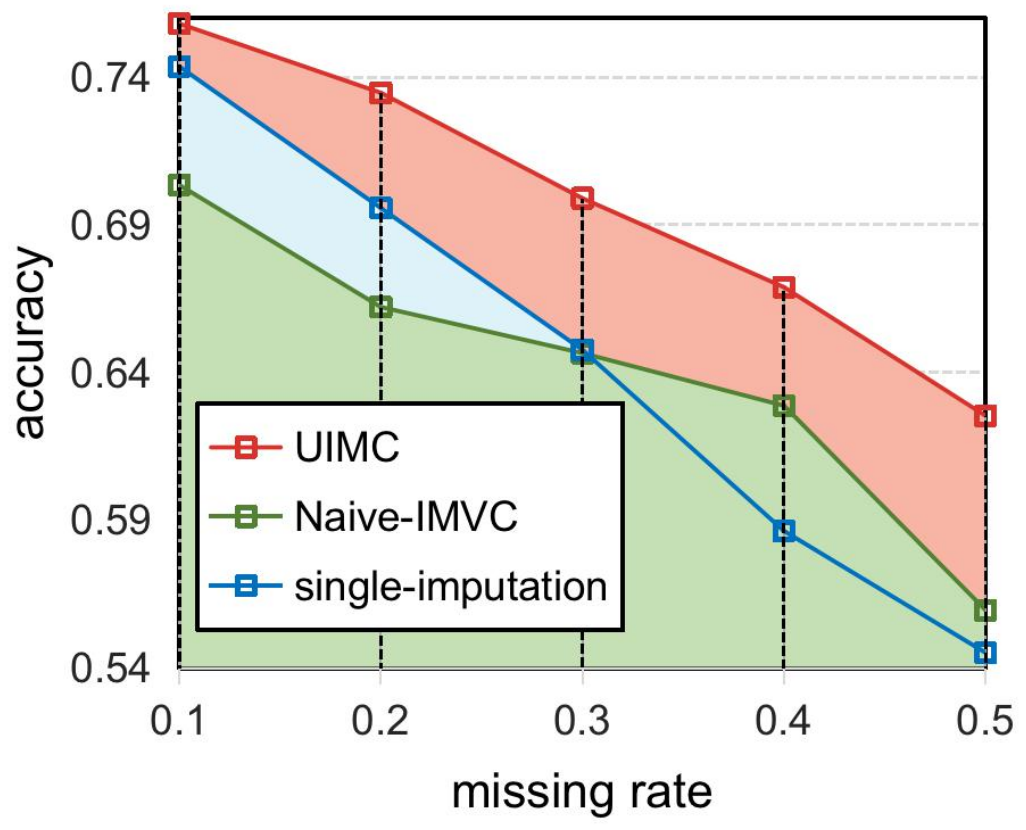}
         \caption{Scene15}
     \end{subfigure}
        \caption{Classification performance of UIMC, single-imputation, and Naive-IMVC on three datasets with $\eta = [0, 0.1, 0.2, 0.3, 0.4, 0.5]$.}
        \label{ablation}
\end{figure*}

\textbf{Q3 Ablation study (I).} To demonstrate the necessity of conducting multiple samplings to impute missing views instead of deterministic imputation, we conduct ablation experiment to compare the multiple samplings with naive deterministic imputation. As shown in Fig.~\ref{ablation},  the method termed single-imputation is designed as imputing the missing views with the mean of neighbors. The superior performance of UIMC verifies that our multiple-imputation strategy can achieve more excellent performance than single-imputation empirically.

\begin{figure*}[htb]
     \centering
     \begin{subfigure}[b]{0.49\textwidth}
         \centering
         \includegraphics[width=\textwidth]{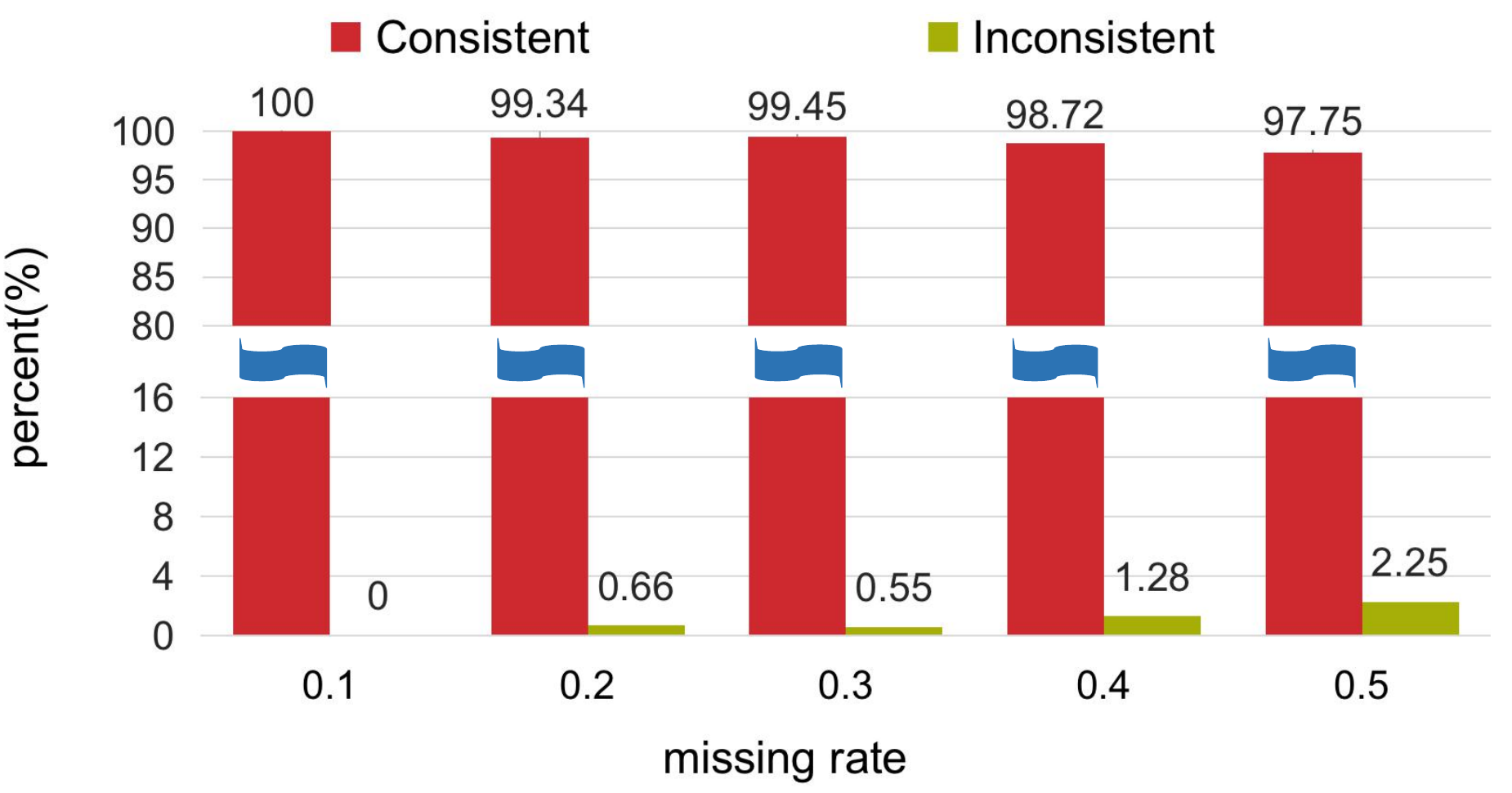}
         \caption{Handwritten}
     \end{subfigure}
     \hfill
     \begin{subfigure}[b]{0.49\textwidth}
         \centering
         \includegraphics[width=\textwidth]{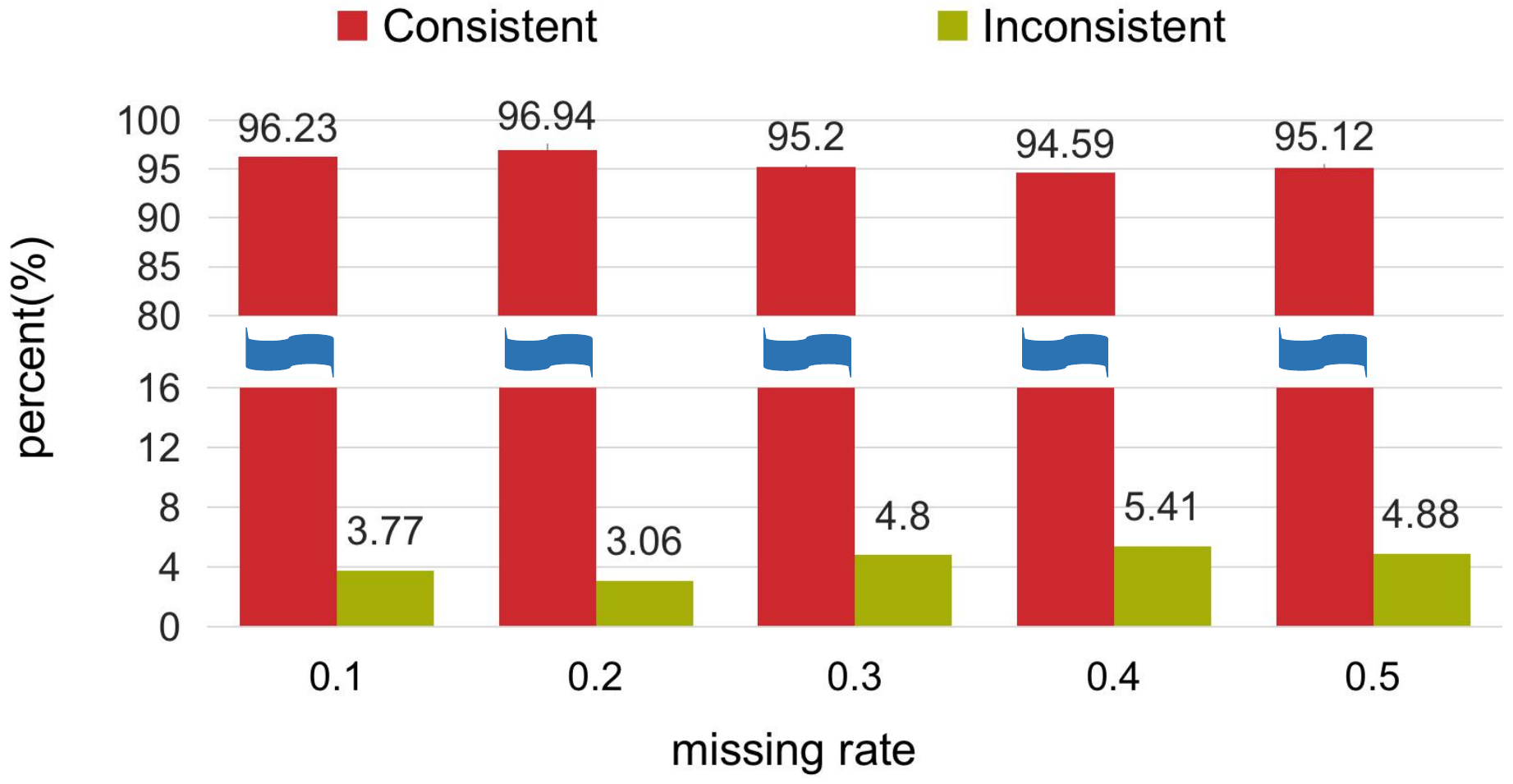}
         \caption{BRCA}
     \end{subfigure}
     \hfill
     \begin{subfigure}[b]{0.49\textwidth}
         \centering
         \includegraphics[width=\textwidth]{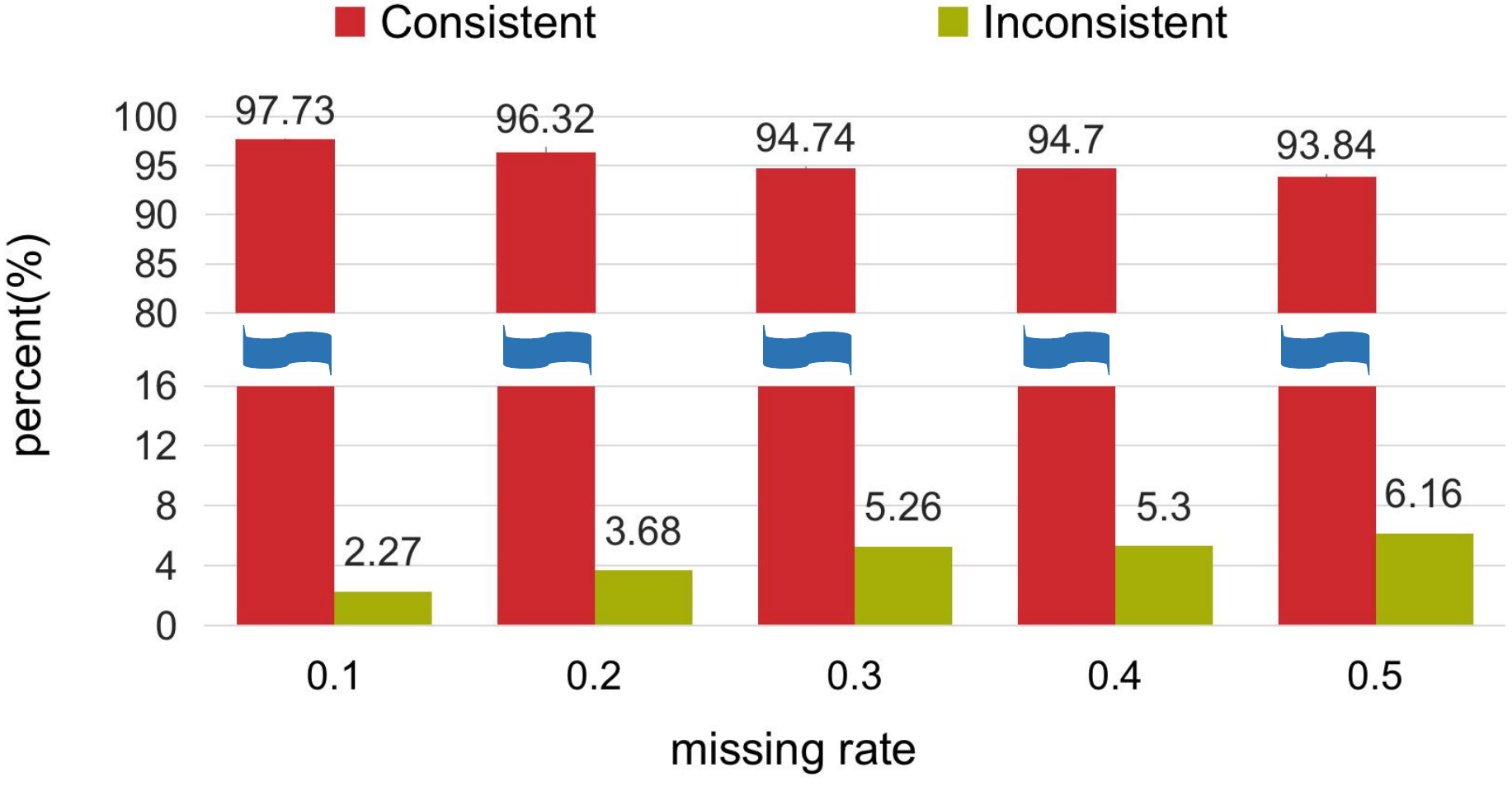}
         \caption{Scene15}
     \end{subfigure}
     \hfill
     \begin{subfigure}[b]{0.49\textwidth}
         \centering
         \includegraphics[width=\textwidth]{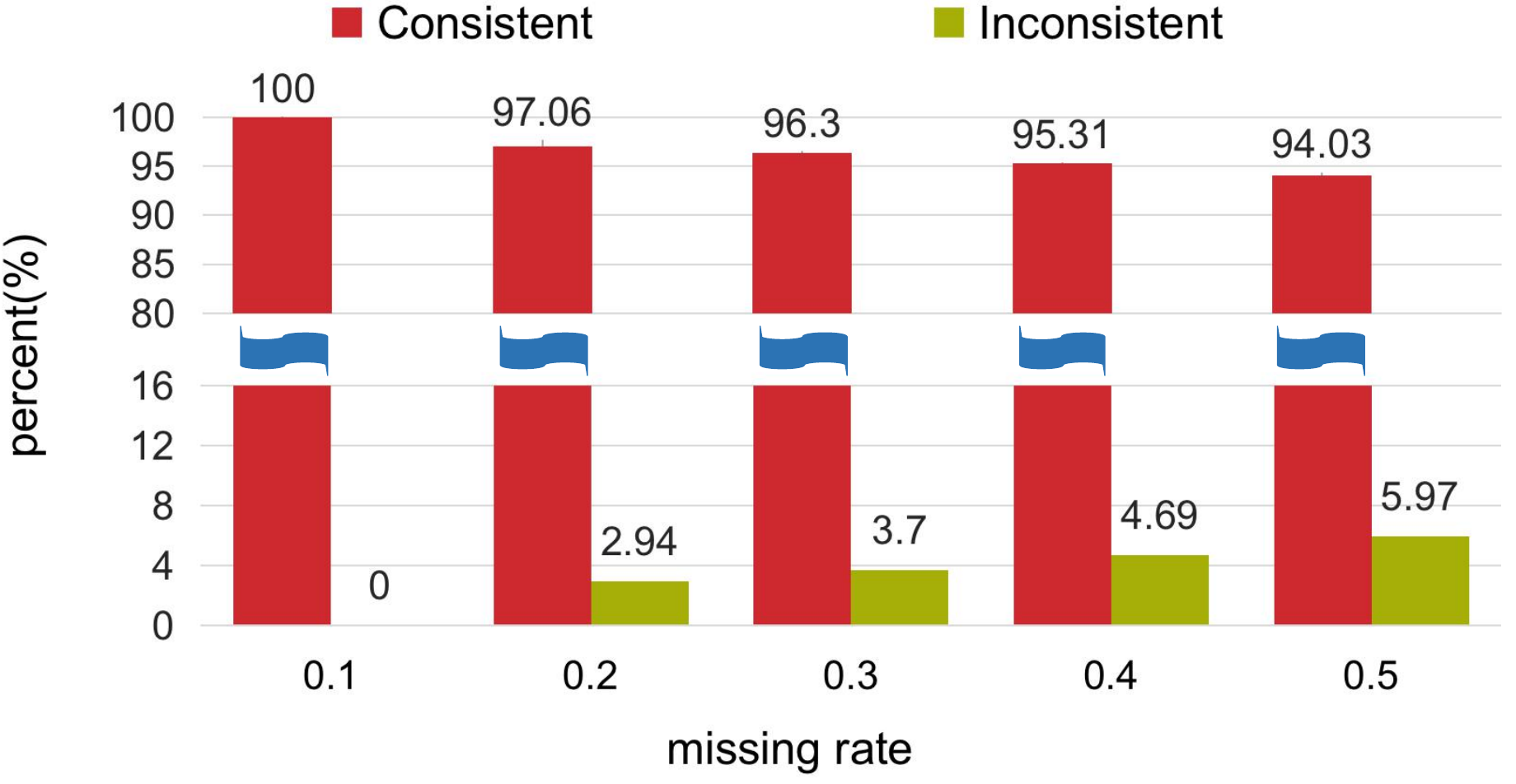}
         \caption{ROSMAP}
     \end{subfigure}
    
\caption{Evaluating the stability of UIMC on four datasets with $\eta = [0, 0.1, 0.2, 0.3, 0.4, 0.5]$.}
\label{stable}
\end{figure*}

\textbf{Q4 Ablation study (II).} We conduct ablation experiment to investigate the effect of exploiting uncertainty in guiding multi-view fusion and classification. Specifically, we compare ours with a naive version termed Naive-IMVC which ignores the uncertainty introduced with imputation. Concretely, the key difference between Naive-IMVC and UIMC is that Naive-IMVC trains classifiers on each single-view and the integrated multi-view data optimized with cross-entropy loss $\mathcal{L}_{ce}=-\sum_{k=1}^K y_{k} \log \left(p_{k}\right)$, where $p_k$ is the predicted probability for class $k$, while UIMC adopts an uncertainty-guided multi-view fusion and classification strategy. As shown in Fig.~\ref{ablation}, we compare UIMC with Naive-IMVC on Handwritten, YaleB, and Scene15 with $\eta = [0, 0.1, 0.2, 0.3, 0.4, 0.5]$. The classification accuracy of UIMC is superior to Naive-IMVC, which indicates that our method can exploit the uncertainty of imputations and make predictions more efficiently and credibly.

\textbf{Q5 Stability. }
We impute missing views by taking multiple samplings from the estimated multivariate Gaussian distributions. 
Since the randomness is involved in sampling process, each sampling operation yields different complete data and then the corresponding classification predictions might be different. To obtain stable and robust prediction, our method adopts a voting strategy to obtain the final classification, which can mitigate the negative impact of the uncertainty from sampling operation. To investigate the prediction variance from the sampling process, we fix the missing patterns of incomplete multi-view data (i.e., the missing views of each sample is fixed), and then conduct $N_i = 10$ samplings to obtain 10 sets of different complete multi-view data. Then we test these 10 imputed multi-view data sets on a pre-trained classification model to obtain 10 prediction results. We conduct this experiment on ROSMAP, BRCA, Handwritten, and Scene15 whose number of classes are 2, 5, 10 and 15, respectively. As shown in Fig.~\ref{stable}, the relatively small proportion of ``Inconsistent'' indicates that the prediction of our model is quite stable, where ``Consistent'' and ``Inconsistent'' indicate the conditions of 10 tests with the same and different predictions, respectively, .

\section{Conclusion}
In this work, we propose a novel framwork for incomplete multi-view classification termed Uncertainty-induced Incomplete
Multi-View Data Classification (UIMC), which can elegantly explore and exploit the uncertainty arising from imputation, producing effectiveness and trustworthiness. Specifically, UIMC focuses on imputing missing views in consideration of the imputation quality. Since UIMC conducts sampling multiple times from the estimated distribution of missing view, it inevitably introduces uncertainty. We employ evidential classifier to characterize the view-specific uncertainty and further utilize Dempster’s combination rule to fuse the uncertain opinions of multiple (imputed) views. We conduct extensive experiments and the empirical results solidly validate the effectiveness and stability of the proposed UIMC. The limitation of this work is that the imputation distributions have been determined before multi-view fusion and classification. Thus, our future work will focus on jointly learning the parameterized distributions of the missing views and the multi-view classifier, and in this way the two components could promote mutually. Another interesting line is theoretical analysis to show the necessity of introducing uncertainty for missing views. 

\section*{Acknowledgement}
This work was supported in part by the National Natural Science Foundation of China (61976151). The authors gratefully acknowledge the support of MindSpore, CANN (Compute Architecture for Neural Networks), and Ascend AI Processor used for this research. We also want to use UIMC on MindSpore, which is a new deep learning computing framework. These problems are left for future work.

%%%%%%%%% REFERENCES
{\small
\bibliographystyle{unsrt}
\bibliography{ref}
}
%\appendix
%\section{Appendix}

\end{document}

%%%%%%%%% TITLE
\title{Supplementary Material for \\ Exploring and Exploiting Uncertainty for Incomplete Multi-View Classification}
\author{
    \IEEEauthorblockN{Mengyao Xie$^{\ddagger}$, Zongbo Han$^{\ddagger}$, Changqing Zhang$^*$, Yichen Bai, Qinghua Hu}\\
    \IEEEauthorblockA{College of Intelligence and Computing, Tianjin University}\\
    \IEEEauthorblockA{\tt\small zhangchangqing@tju.edu.cn}
}
\maketitle

\thispagestyle{empty}

\section{Large-Scale Multi-view Datasets Experiments}
To validate the effectiveness of the proposed method on large-scale datasets, we conduct experiments on two datasets.
\textbf{Caltech-101}~\cite{Caltech} is a 6-view dataset consists of pictures of objects belonging to 101 classes, plus one background clutter class. Each class contains roughly 40 to 800 images, totalling 9144 images. 
\textbf{NUS-WIDE-OBJECT}~\cite{NUS} is a 5-view dataset consists of pictures belonging to 31 classes, totally 30,000 images.

\begin{figure}[!htb]
     \centering
     \begin{subfigure}[b]{0.49\textwidth}
         \centering
         \includegraphics[width=\textwidth]{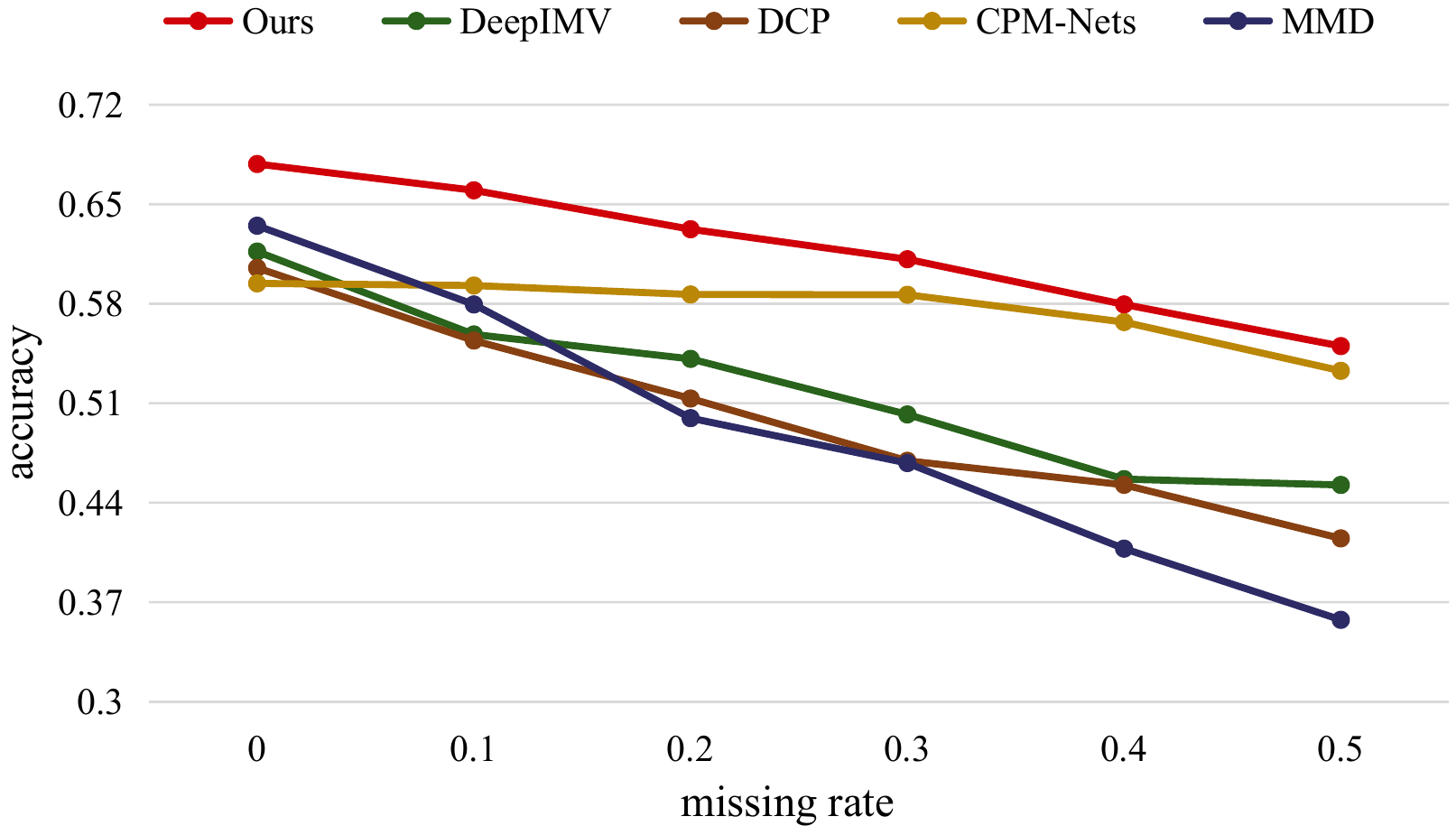}
         \caption{Caltech-101 of 9144 samples}
     \end{subfigure}
     \hfill
     \begin{subfigure}[b]{0.49\textwidth}
         \centering
         \includegraphics[width=\textwidth]{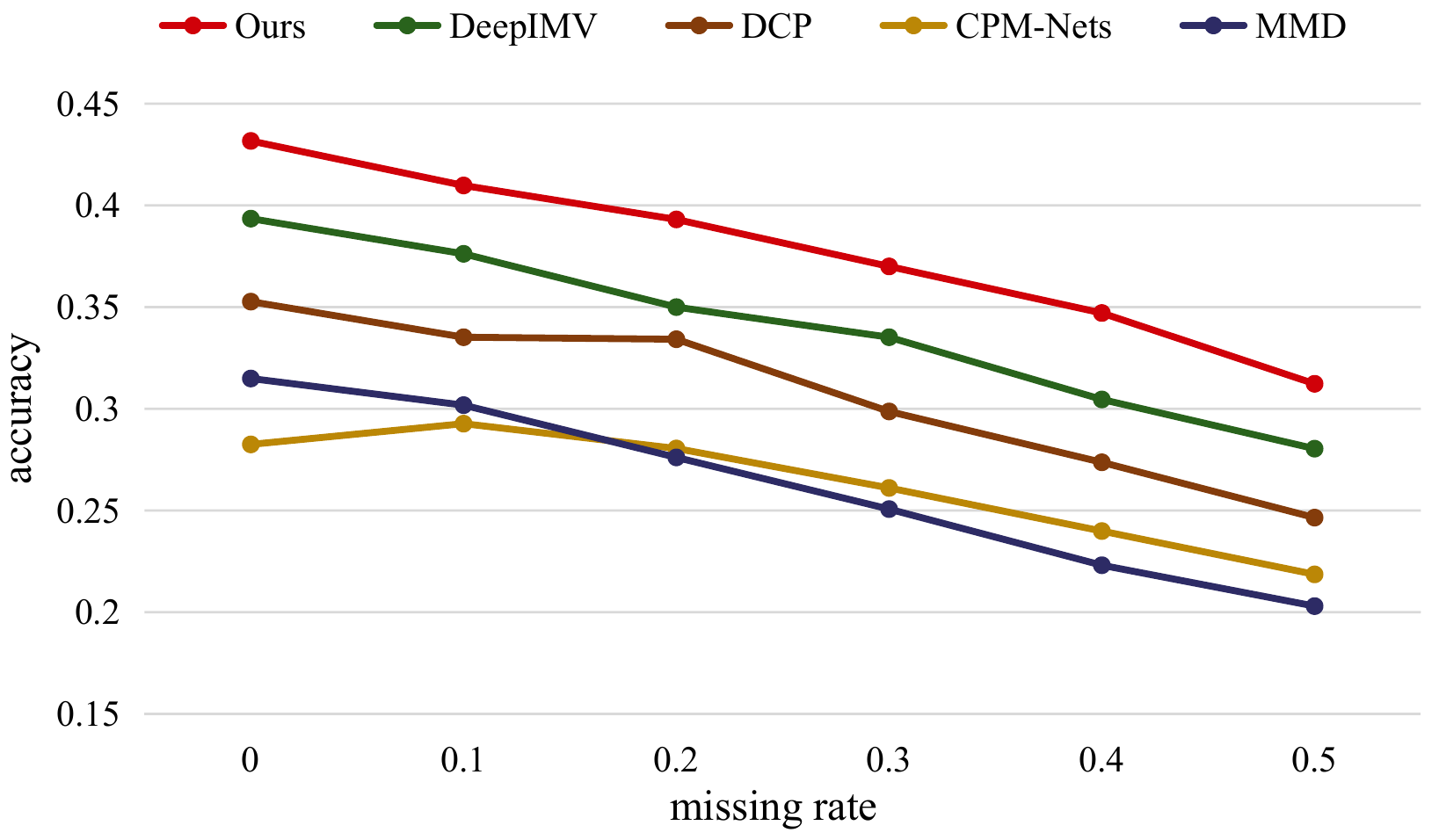}
         \caption{NUS-WIDE-OBJECT of 30,000 samples}
     \end{subfigure}
     \caption{Classification performance with $\eta = [0, 0.1, 0.2, 0.3, 0.4, 0.5]$.}
     \label{acc}
\end{figure}

We compare the proposed algorithm with 4 state-of-the-art multi-view classification methods.
\textbf{CPM-Nets}~\cite{CPM} directly learns the joint latent representations for all views with available data, and maps the latent representation to classification predictions.
\textbf{DeepIMV}~\cite{DeepIMV} applies the information bottleneck (IB) framework to obtain marginal and joint representations with the available data, and constructs the view-specific and multi-view predictors to obtain the classification predictions.
\textbf{MMD}~\cite{MMD} models both the feature-level and modality-level dynamicities and introduces a sparse gating strategy to trustworthily fuse the complete multi-view data.
\textbf{DCP}~\cite{DCP} provides an information theoretical framework under which the consistency learning and data recovery are treated as a whole. The missing views are recovered by minimizing the conditional entropy through dual prediction. 
For the MMD method which can not deal with incomplete multi-view data, we first impute the missing views with the corresponding means, and then train MMD to obtain the classification predictions.  
As shown in Fig.~\ref{acc}, our method still achieves the superior performance on large-scale datasets.

\section{Details and Impact of Hypeparameter $\lambda$}
\begin{figure}[!htb]
     \centering
     \begin{subfigure}[b]{0.49\textwidth}
         \centering
         \includegraphics[width=\textwidth]{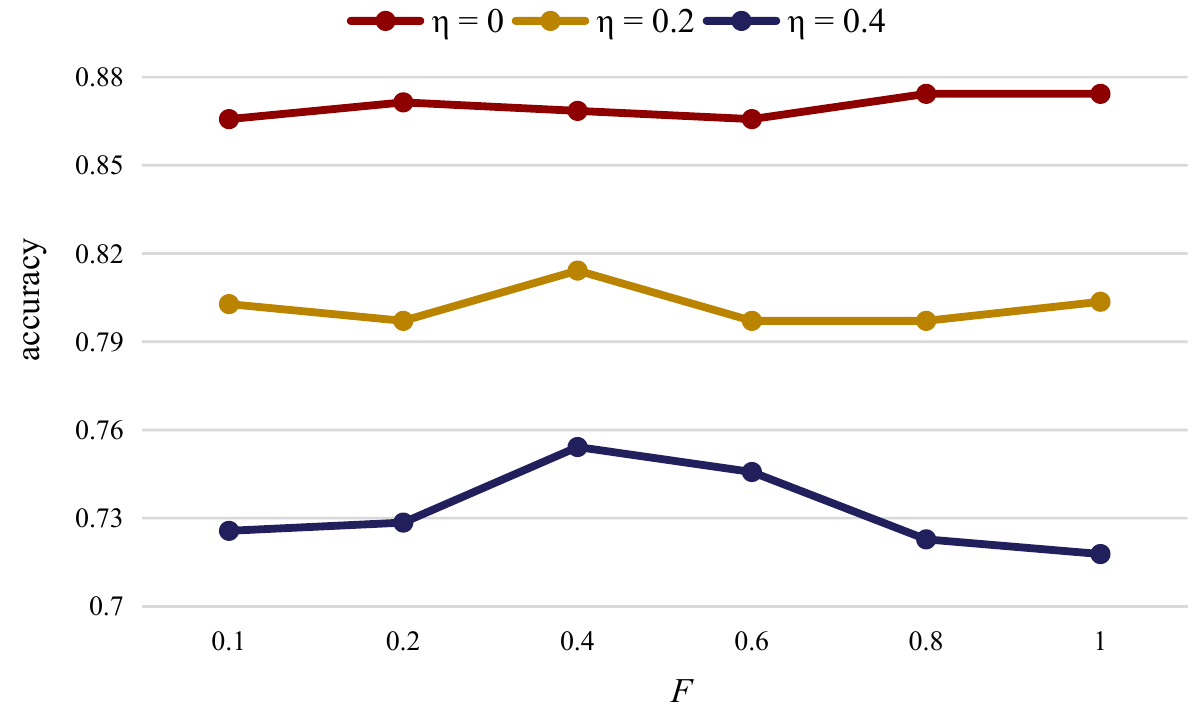}
         \caption{Final value}
         \label{parameter1}
     \end{subfigure}
     \hfill
     \begin{subfigure}[b]{0.49\textwidth}
         \centering
         \includegraphics[width=\textwidth]{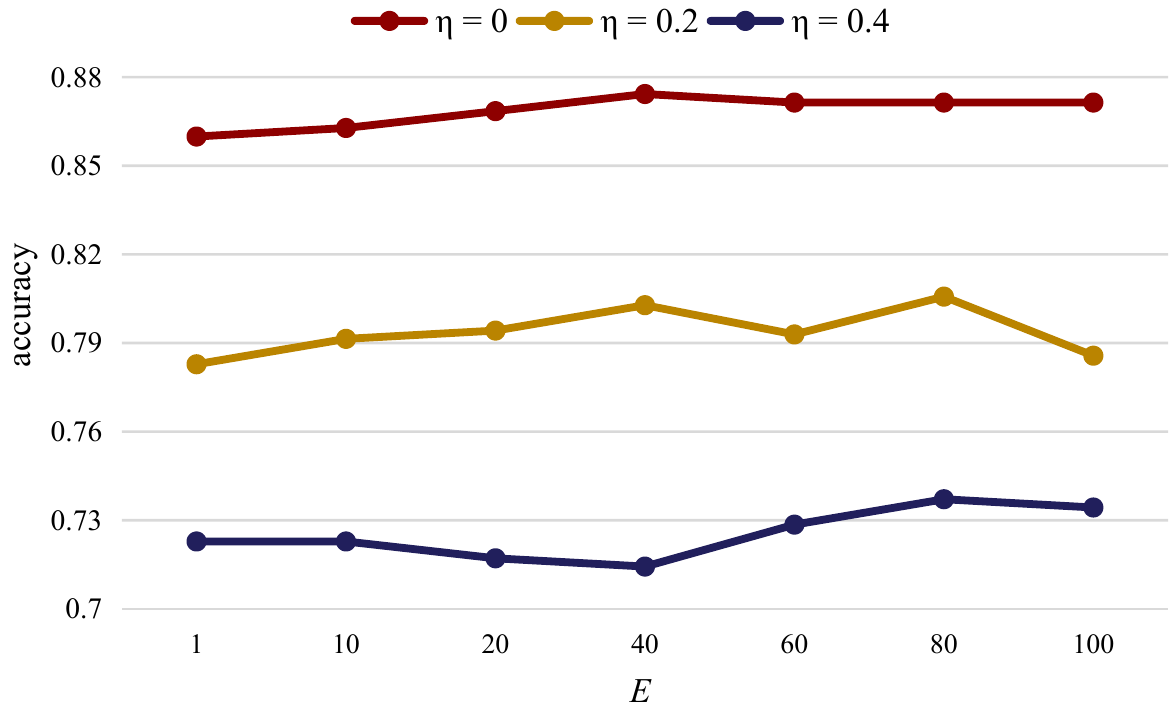}
         \caption{Decay rate}
         \label{parameter2}
     \end{subfigure}
     \caption{Classification performance with different final values and decay rates of $\lambda$ on ROSMAP under $\eta = [0, 0.2, 0.4]$ where we set $E = 50$ in Fig.~\ref{parameter1} and $F = 1$ in Fig.~\ref{parameter2}. }
     \label{parameter}
\end{figure}

\textbf{\textit{Details.}} Same as previous methods~\cite{sensoy2018evidential,han2022trusted}, we set $\lambda = min(F, e/E)$ where $F, e$ and $E$ denote the final value of $\lambda$, iteration epoch and the parameter controlling the decay rate of $\lambda$. \textbf{\textit{Impact.}} we conduct experiments on ROSMAP with different missing rates $\eta$ to investigate the influence of $\lambda$ in terms of decay rate and the final value. Specifically, we change the final value $F$ or parameter $E$ to control the decay rate. As shown in Fig.~\ref{parameter}, our method is robust to the final value and decay rate of $\lambda$.

\section{Details of Constructing the Incomplete Multi-view Data.}
Same as previous methods~\cite{CPM, DeepIMV}, we construct the incomplete multi-view data to satisfy the following requirements. 1. Missing at random with the desired missing rate $\eta$. 2. Each sample has at least one view that is not missing. Therefore, the following steps are conducted. 1. Calculate how many views are missing with $N^m = \eta N V$, where $N$ and $V$ are number of samples and views respectively. 2. Randomly assign $N^m$ to each sample while ensure that at least one view is available for each sample (i.e., $N^m=\sum_{i=1}^N N_i^m$ and $N_i^m<V$, where $N_i^m$ is the number assigned to each sample). 3. Randomly select $V-N_i^m$ available views for each sample.

{\small
\bibliographystyle{unsrt}
\bibliography{ref}
}